\documentclass[]{opendatalab}
\usepackage{caption}
\usepackage{xcolor}
\usepackage{longtable}
\usepackage{listings}
\usepackage[numbers]{natbib}
\usepackage{amssymb}
\usepackage{tabularx}
\usepackage{float}
\usepackage{graphicx}
\usepackage{makecell}
\usepackage[table]{xcolor}
\usepackage{cleveref}
\usepackage{array}
\usepackage[misc]{ifsym}

\makeatletter
\newcommand\andauthor{%
  \g@addto@macro\authorlist{\par\vspace{1mm}}
}
\newcommand\nlauthor[2][]{%
  \addtolist[#1]{#2}{\authorlist}{\authorformat}{}
}
\makeatother

\RequirePackage{xspace}
\makeatletter
\DeclareRobustCommand\onedot{\futurelet\@let@token\@onedot}
\def\@onedot{\ifx\@let@token.\else.\null\fi\xspace}

\def\eg{\emph{e.g}\onedot}

\def\etc{\emph{etc}\onedot} \def\vs{\emph{vs}\onedot}

\makeatother
\newcommand{\mineru}{MinerU2.5\xspace}
\newcommand{\minerupro}{MinerU2.5-Pro\xspace}

\definecolor{codegreen}{rgb}{0,0.6,0}
\definecolor{codegray}{rgb}{0.5,0.5,0.5}
\definecolor{codepurple}{rgb}{0.58,0,0.82}
\definecolor{backcolour}{rgb}{0.95,0.95,0.92}
\definecolor{promptcolor}{HTML}{D1D0F2}
\definecolor{promptcolorheader}{HTML}{bdbcec}
\newcommand{\promptbox}[2]{
\begin{tcolorbox}[
top=0.3em,bottom=0.3em,left=0.5em,right=0.5em,
toptitle=0.3em,bottomtitle=0.2em,boxsep=0pt,
colframe=promptcolorheader,colback=promptcolor!50,boxrule=0.5pt,
]
\footnotesize
\end{tcolorbox}
}
\lstdefinestyle{mystyle}{
    backgroundcolor=\color{backcolour},   
    commentstyle=\color{codegreen},
    keywordstyle=\color{magenta},
    numberstyle=\tiny\color{codegray},
    stringstyle=\color{codepurple},
    basicstyle=\ttfamily\footnotesize,
    breakatwhitespace=false,         
    breaklines=true,                 
    captionpos=b,                    
    keepspaces=true,                 
    numbers=left,                    
    numbersep=5pt,                  
    showspaces=false,                
    showstringspaces=false,
    showtabs=false,                  
    tabsize=2
}

\title{MinerU2.5-Pro: Pushing the Limits of Data-Centric \mbox{Document Parsing at Scale}}

\vspace{10pt}

\author[1*\ddagger]{\quad \quad \quad Bin Wang}
\author[1*]{Tianyao He}
\author[1*]{Linke Ouyang}
\author[1*]{Fan Wu}
\author[1*]{Zhiyuan Zhao}
\author[1*]{Tao Chu}
\andauthor
\nlauthor[1*]{\quad \quad Yuan Qu}
\author[1*]{Zhenjiang Jin}
\author[1*]{Weijun Zeng}
\author[1*]{Ziyang Miao}
\author[1*]{Bangrui Xu}
\author[1,2*]{Junbo Niu}
\andauthor
\nlauthor[1]{\;Mengzhang Cai}
\author[1]{Jiantao Qiu}
\author[1,2]{Qintong Zhang}
\author[1]{Dongsheng Ma}
\author[1]{Yuefeng Sun}
\author[1]{Hejun Dong}
\andauthor
\nlauthor[1,2]{\ Wenzheng Zhang}
\author[1]{Jutao Xiao}
\author[1]{Jiayong Shi}
\author[1]{Pengyu Liao}
\author[1]{Xiaomeng Zhao}
\author[4]{Huaping Zhong}
\andauthor
\nlauthor[1]{\quad \quad Liqun Wei}
\author[1]{Jing Yu}
\author[1]{Jie Yang}
\author[1]{Wei Li}
\author[1]{Shasha Wang}
\author[1]{\;Qianqian Wu}
\author[1,3]{\;Xuanhe Zhou}
\andauthor
\nlauthor[1]{\quad \quad Weijia Li}
\author[1]{Zhenxiang Li}
\author[1]{Zhongying Tu}
\author[1]{Jiang Wu}
\author[1]{Lijun Wu}
\author[1]{Chao Xu}
\author[1]{Kai Chen}
\andauthor
\nlauthor[1,2]{\quad \quad  \quad  \quad Wentao Zhang}
\author[1]{Yu Qiao}
\author[1]{Bowen Zhou}
\author[1\ \textrm{\Letter}]{Dahua Lin}
\author[1\ \textrm{\Letter}]{Conghui He}

\affiliation[1]{Shanghai Artificial Intelligence Laboratory}
\affiliation[2]{Peking University}
\affiliation[3]{Shanghai Jiao Tong University}
\affiliation[4]{SenseTime}

\abstract{

Current document parsing methods advance primarily through model architecture innovation, while systematic engineering of training data remains underexplored. Yet state-of-the-art models spanning diverse architectures and parameter scales exhibit highly consistent failure patterns on the same set of hard samples, suggesting that the performance bottleneck stems from shared deficiencies in training data rather than from architectural differences. Building on this finding, we present \minerupro, which advances the state of the art purely through data engineering and training strategy design while retaining the 1.2B-parameter architecture of \mineru unchanged. At its core is a Data Engine co-designed around coverage, informativeness, and annotation accuracy: Diversity-and-Difficulty-Aware Sampling expands training data from under 10M to 65.5M samples while mitigating distribution shift; Cross-Model Consistency Verification leverages output consensus among heterogeneous models to assess sample difficulty and generate reliable annotations; the Judge-and-Refine pipeline improves annotation quality for hard samples through render-then-verify iterative correction. A three-stage progressive training strategy---large-scale pre-training, hard sample fine-tuning, and GRPO alignment---sequentially exploits these data at different quality tiers. On the evaluation front, we rectify element-matching biases in OmniDocBench~v1.5 and introduce a Hard subset, establishing the more discriminative OmniDocBench~v1.6 protocol. Without any architectural modification, \minerupro achieves 95.69 on OmniDocBench~v1.6, improving over the same-architecture baseline by 2.71 points and surpassing all existing methods, including those based on models with over 200$\times$ more parameters.

}

\metadata[* Equal contribution\quad $\textrm{\Letter}$ Corresponding author \quad $\ddagger$ Project leader]{}
\correspondence{Conghui He, \email{heconghui@pjlab.org.cn}}
\metadata[Code:]{\url{https://github.com/opendatalab/MinerU}}
\metadata[Model:]{\url{https://huggingface.co/opendatalab/MinerU2.5-Pro-2604-1.2B}}
\date{\today}

\begin{document}

\maketitle

\newpage
\tableofcontents
\newpage
\section{Introduction}
\label{sec:introduction}

\begin{figure*}[t]
\centering
\includegraphics[width=\textwidth]{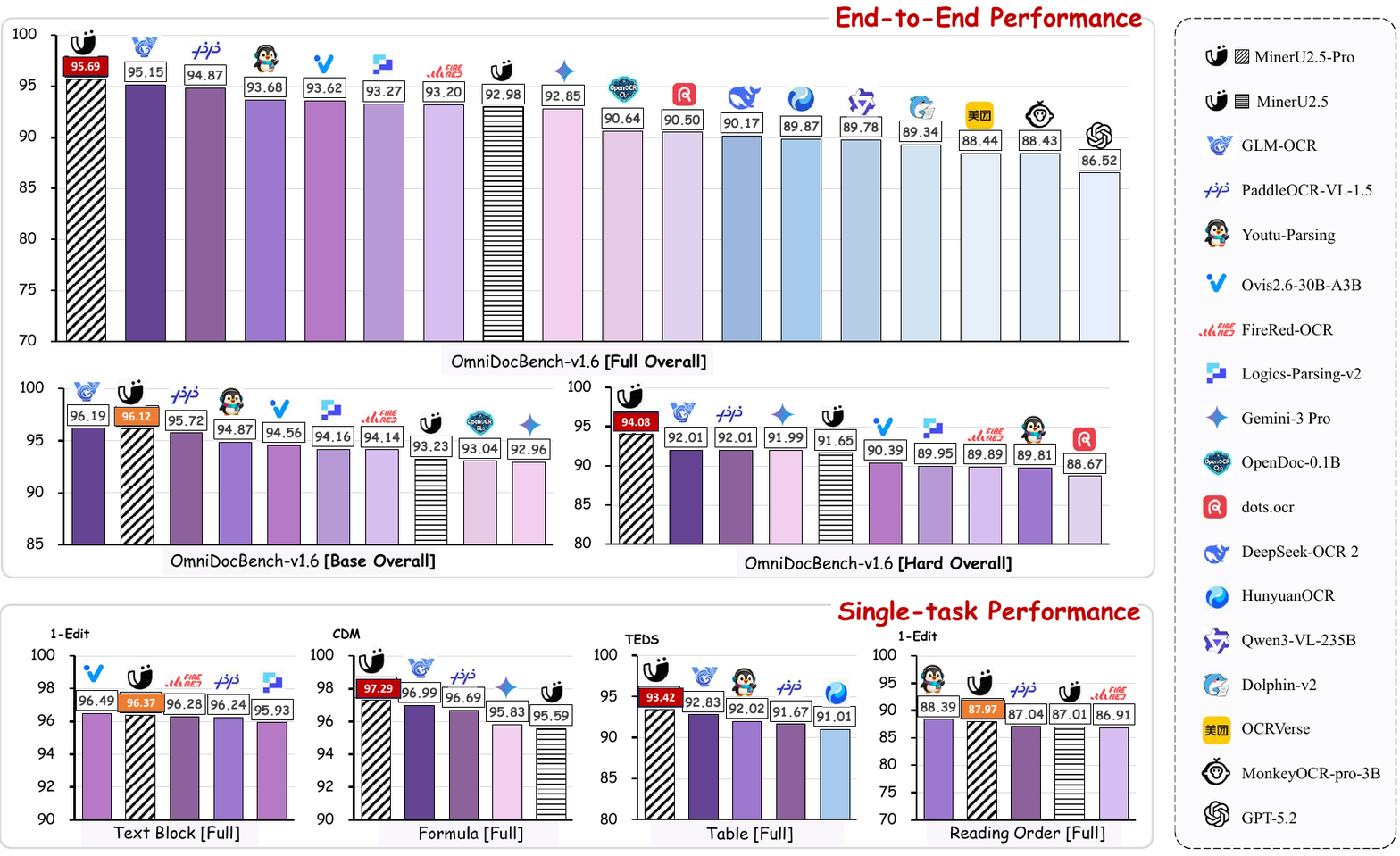}
\caption{Performance comparison on OmniDocBench~v1.6, which comprises Base (standard samples), Hard (challenging samples), and Full (overall) splits. Built upon \mineru~\cite{mineru25} with its 1.2B-parameter architecture entirely unchanged, \minerupro improves the overall score from 92.98 to 95.69 purely through data engineering and training strategy design, outperforming both specialized document parsing models (\eg GLM-OCR~\cite{glm_ocr}, PaddleOCR-VL-1.5~\cite{paddleocr_vl15}, Youtu-Parsing\cite{yin2026youtuparsingperceptionstructuringrecognition}) and general-purpose VLMs (\eg Gemini~3~Pro, Qwen3-VL-235B~\cite{yang2025qwen3technicalreport}). Detailed results are presented in \Cref{tab:main_results}.}
\label{fig:leaderboard}
\end{figure*}

Document parsing converts unstructured documents such as PDFs into structured, machine-readable formats (\eg Markdown), serving as critical infrastructure for LLM training data pipelines~\cite{zhang2024document,olmocr,mineru} and retrieval-augmented generation systems~\cite{rag_anything,bookrag,zhang2025ocr}. As end-to-end approaches based on vision-language models (VLMs) progressively replace traditional pipeline systems~\cite{nougat,got_ocr,mineru25}, research has focused predominantly on architectural innovation and inference efficiency, leading to rapid score convergence among top models on standard benchmarks.

Yet this convergence raises a deeper question: what constitutes the remaining performance bottleneck? Our cross-analysis of parsing results from multiple state-of-the-art models---spanning diverse architectures and parameter scales---on large-scale real-world PDFs reveals a striking pattern: these models exhibit highly similar failure modes on the same hard samples, with certain parsing errors common to all tested systems. Since these systematic failures transcend any particular architecture, they point to a common root cause: the current performance bottleneck in document parsing stems primarily from shared deficiencies in training data, not from model architecture itself.

This data bottleneck manifests in two interrelated dimensions. First, \emph{insufficient coverage}: for instance, \mineru's training data totals less than 10M pages with distributions concentrated on high-frequency categories, severely underrepresenting long-tail scenarios such as complex nested tables and dense formula layouts. Second, an \emph{annotation quality paradox}: the hard samples that contribute most to model improvement are precisely those for which automatic annotation is least reliable, since no mainstream model can consistently parse them correctly. Structural annotations for complex tables and \LaTeX{} transcriptions of dense formulas are highly error-prone, and this annotation noise propagates directly into model behavior during supervised fine-tuning. These two issues are deeply intertwined: simply scaling data volume is insufficient to raise the performance ceiling, as added data only amplifies existing distribution biases and annotation noise.

Beyond data, our cross-analysis also exposes blind spots in the existing evaluation framework. OmniDocBench~v1.5 contains relatively few hard samples, and its element-matching logic exhibits systematic biases toward specific output formats, introducing scoring artifacts that complicate fair cross-system comparison. We accordingly introduce OmniDocBench~v1.6, which corrects these matching biases and incorporates a dedicated Hard subset to establish a Base/Hard/Full three-tier evaluation protocol.

Based on this analysis, we argue that as model architectures mature, systematic data engineering becomes the primary lever for advancing document parsing performance. To test this hypothesis, we build \minerupro---retaining the identical 1.2B-parameter decoupled coarse-to-fine architecture of \mineru~\cite{mineru25} and focusing all optimization on the Data Engine and training strategy, ensuring that all performance gains are attributable to data-level improvements. On OmniDocBench~v1.6, \minerupro achieves 95.69 (baseline 92.98, +2.71), surpassing all existing methods, including models with over 200$\times$ more parameters (\Cref{fig:leaderboard}).

Our contributions are summarized as follows:

\begin{itemize}
    \item A Data Engine co-designed around coverage, informativeness, and annotation accuracy. It comprises three core components---Diversity-and-Difficulty-Aware Sampling (DDAS), Cross-Model Consistency Verification (CMCV), and a Judge-and-Refine annotation pipeline---that together expand training data from under 10M to 65.5M pages while systematically improving annotation quality through a closed-loop progression from sampling to refinement.

    \item A three-stage progressive training strategy---large-scale pre-training, high-quality hard sample fine-tuning, and GRPO format alignment---matched to the data quality tiers produced by the Data Engine. With these data and training improvements alone, the same 1.2B-parameter model achieves state-of-the-art performance on OmniDocBench~v1.6, surpassing all existing methods.

    \item OmniDocBench~v1.6, an upgraded evaluation protocol that corrects element-matching biases in v1.5 through Multi-Granularity Adaptive Matching and introduces a Hard subset, establishing a Base/Hard/Full three-tier framework for fairer and more discriminative evaluation.
\end{itemize}

\section{Related Work}
\label{sec:related_work}

\subsection{Document Parsing Methods}
\label{subsec:doc_parsing_methods}

Existing document parsing methods fall into three paradigms based on system architecture.

\paragraph{Pipeline-based methods.}
These methods decompose document parsing into independent subtasks---layout detection, text recognition, table extraction, formula recognition---and execute them in a cascade~\cite{mineru,zhao2024doclayout,livathinos2025,marker,cui2025paddleocr}. This modular design enables independent optimization of each component but suffers from error propagation and inter-module information loss.

\paragraph{End-to-end VLM methods.}
These methods directly map document images to structured output, avoiding the cascading errors inherent in pipeline approaches. Nougat~\cite{nougat}, built on the Donut architecture~\cite{donut}, established a strong baseline for the image-to-markup paradigm on academic documents; GOT-OCR~2.0~\cite{got_ocr} unified scene text and document OCR within a single model. Subsequent works such as Ocean-OCR~\cite{ocean_ocr}, olmOCR~\cite{olmocr}, and dots.ocr~\cite{dots_ocr} employ native-resolution visual encoders to further improve performance. However, native-resolution processing incurs $\mathcal{O}(N^2)$ token complexity, creating efficiency bottlenecks for high-resolution documents.

\paragraph{Decoupled VLM methods.}
These methods separate layout analysis from content recognition, combining the controllability of pipeline approaches with the semantic modeling power of VLMs. Early works such as Dolphin~\cite{dolphin} and MonkeyOCR~\cite{monkeyocr} demonstrated the viability of this paradigm but faced limitations in resolution handling or system complexity. \mineru~\cite{mineru25} unifies layout analysis and content recognition within a single 1.2B-parameter model with native-resolution support~\cite{nativeres}, balancing resolution fidelity, efficiency, and deployment complexity. Subsequent works extend the decoupled paradigm along various axes: multi-token prediction for throughput~\cite{glm_ocr}, diffusion-based decoding for improved parsing efficiency and robustness~\cite{minerudiffusion}, non-planar document handling~\cite{paddleocr_vl15}, in-the-wild robustness~\cite{hunyuanocr}, and high-compression vision-text mappings~\cite{deepseek_ocr}. General-purpose VLMs such as Gemini~2.5~Pro~\cite{gemini25pro} and Qwen2.5-VL-72B~\cite{qwen25vl} also achieve competitive results, though their large parameter scales hinder cost-effective deployment at production scale.

Across these works, the main line of methodological evolution focuses on architecture design and inference efficiency, while systematic engineering of training data---co-optimizing coverage, informativeness, and annotation accuracy---has not been adequately explored as an independent research problem. Our work addresses this dimension and is largely complementary to the architectural advances above.

\subsection{Data-Centric AI}
\label{subsec:data_centric_ai}

The Data-Centric AI paradigm~\cite{ng2021datacentric,zha2023datacentric} advocates systematically improving data quality while keeping the model fixed, and has been validated in vision-language pretraining~\cite{gadre2023datacomp} and large language model fine-tuning~\cite{zhou2024lima}. In document parsing, however, data engineering remains fragmented: olmOCR~\cite{olmocr} emphasizes data scale expansion over quality stratification; DocGenome~\cite{docgenome} is restricted to academic papers and lacks difficulty differentiation; existing technical reports~\cite{mineru25,glm_ocr,paddleocr_vl15} describe training data but treat it as a prerequisite for model training rather than an independent research subject.

Our work treats data construction for document parsing as a standalone systematic research problem, co-optimizing coverage, informativeness, and annotation accuracy within a unified framework. Methodologically, our CMCV approach draws on the core principles of ensemble-based active learning~\cite{seung1992qbc} and query-by-committee~\cite{freund1997qbc} by leveraging multi-model disagreement to quantify sample informativeness. Beyond standard disagreement-based selection, CMCV couples difficulty information with downstream annotation strategies in a closed loop and addresses the document-parsing-specific challenge of unreliable hard sample annotation through the Judge-and-Refine pipeline.

\subsection{Document Parsing Evaluation}
\label{subsec:doc_parsing_eval}

Document parsing evaluation involves both metric design and evaluation protocol. At the metric level, text recognition commonly uses edit distance~\cite{levenshtein1966}, table structure recovery uses TEDS~\cite{zhong2020teds}, and formula recognition has recently shifted from BLEU to CDM (Character Detection Matching)~\cite{wang2025cdm}. OmniDocBench~\cite{omnidocbench} integrates all three metrics and provides one of the most comprehensive document parsing evaluation frameworks to date; OCRBench~\cite{ocrbench} and CC-OCR~\cite{cc_ocr} focus on evaluating multimodal models' overall OCR capabilities.

At the protocol level, however, the critical impact of element-matching strategies on evaluation fairness remains largely overlooked. End-to-end systems vary in output granularity, segmentation strategies, and format conventions, and the choice of matching algorithm systematically affects evaluation scores. We identify such systematic biases in OmniDocBench~v1.5 and rectify them in v1.6 through Multi-Granularity Adaptive Matching (detailed in \Cref{sec:omnidocbench}).

\section{Data Engine}
\label{sec:data_engine}

To address the data deficiencies identified above, we first examine the limitations of existing data pipelines. \mineru~\cite{mineru25} built a data pipeline comprising cluster-based sampling, Iterative Model Inference Consistency (IMIC) hard sample mining, and model-based annotation refinement, but these components operate independently without joint optimization of coverage, informativeness, and accuracy: sampling is not informed by difficulty, annotation refinement applies a uniform strategy regardless of sample difficulty, and hard samples mined by IMIC still face unreliable automatic annotation. Similar limitations exist in PaddleOCR-VL-1.5's Uncertainty-Aware Cluster Sampling (UACS)~\cite{paddleocr_vl15}.

\begin{figure*}[t]
\centering
\includegraphics[width=\textwidth]{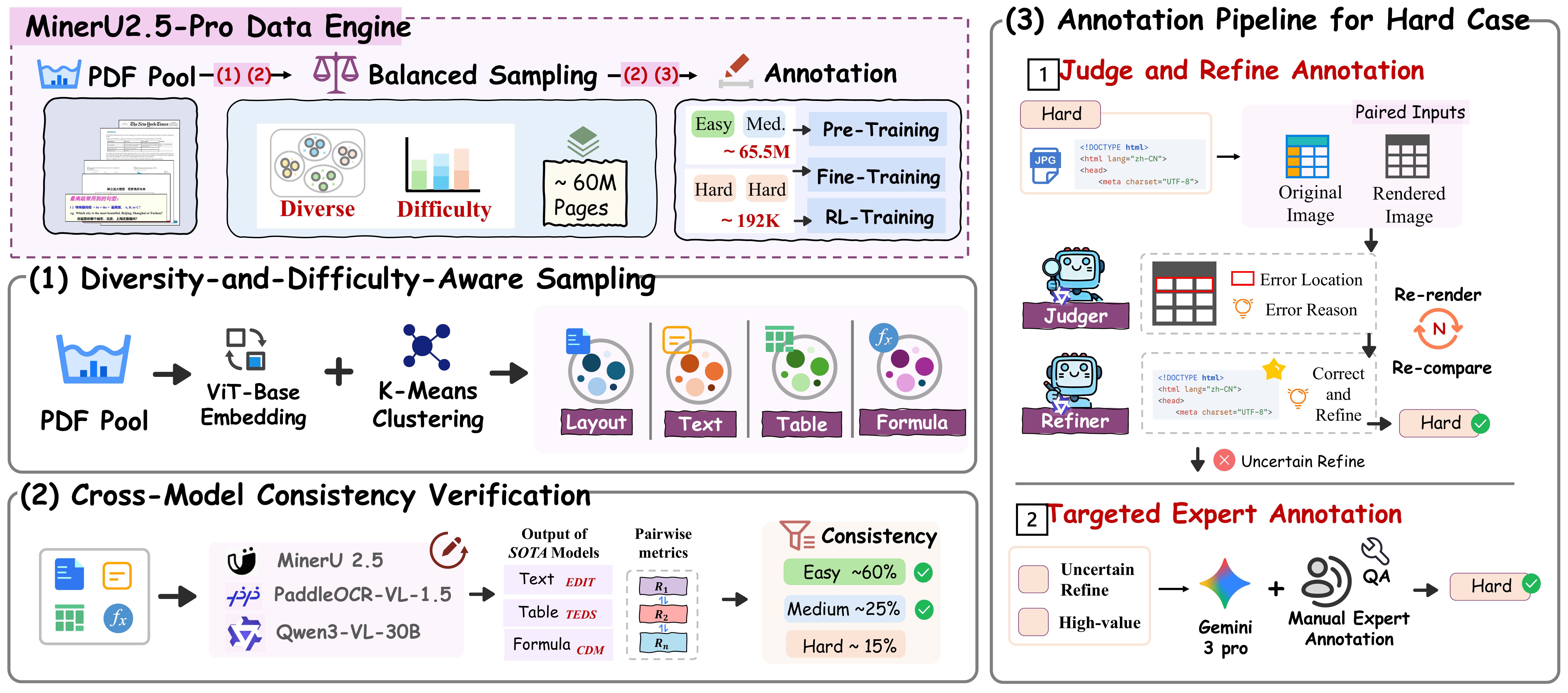}
\caption{Overview of the Data Engine pipeline. The system co-optimizes three dimensions---Coverage, Informativeness, and Accuracy---through three synergistic stages: Diversity-and-Difficulty-Aware Sampling (DDAS), Cross-Model Consistency Verification (CMCV), and the Annotation Pipeline for Hard Case.}
\label{fig:data_engine}
\end{figure*}

The Data Engine of \minerupro is co-designed around these three dimensions. DDAS expands data coverage through task-aware clustering and mitigates distribution shift (\Cref{subsec:ddas}). CMCV performs difficulty stratification on the sampled data via multi-model cross-validation, identifying highly informative samples (\Cref{subsec:cmcv}). The Annotation Pipeline for Hard Case improves annotation accuracy through render-then-verify iterative correction, with residual samples beyond automatic correction routed to targeted expert annotation to guarantee final quality (\Cref{subsec:annotation_pipeline}). Together, these components form a coarse-to-fine quality progression, enabling simultaneous data scaling (under 10M $\to$ 65.5M) and annotation quality improvement. The overall pipeline is illustrated in \Cref{fig:data_engine}.

\subsection{Diversity-and-Difficulty-Aware Data Sampling}
\label{subsec:ddas}

Training data for document parsing exhibits a typical long-tail distribution problem: high-frequency categories (\eg standard academic papers, single-column reports) dominate the data pool, while long-tail scenarios such as complex nested tables, dense formula layouts, and unconventional multi-column layouts are severely underrepresented. As noted above, existing approaches~\cite{mineru25,paddleocr_vl15} rely on single-model signals for difficulty estimation, which cannot distinguish model-specific weaknesses from universally hard samples.

\begin{figure*}[t]
\centering
\includegraphics[width=\textwidth]{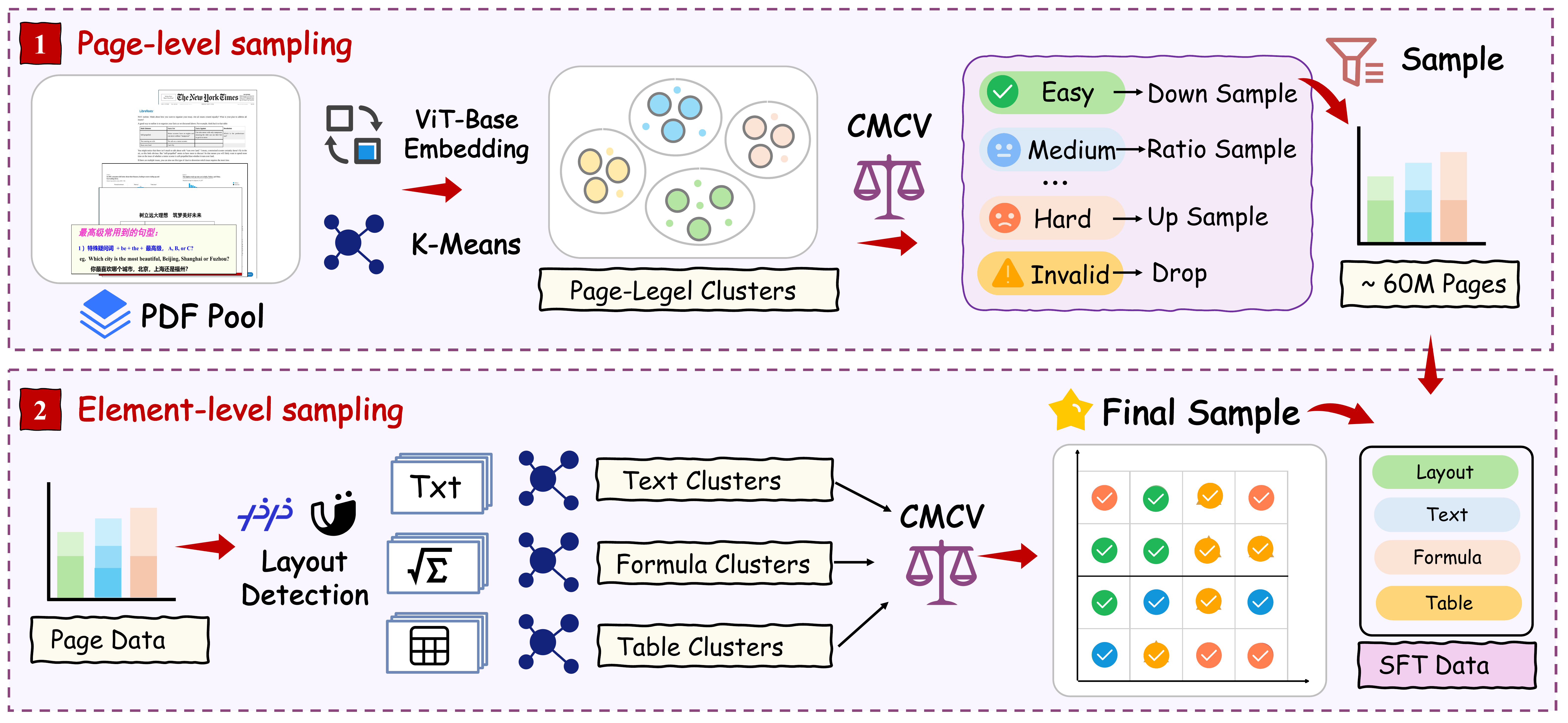}
\caption{The DDAS pipeline operates at two granularity levels. \textbf{Upper}: Page-level sampling for layout detection data---pages from the PDF pool are embedded via ViT-base, clustered, and resampled by jointly weighting cluster diversity and CMCV-derived difficulty, yielding about 60M pages with balanced distribution and difficulty coverage. \textbf{Lower}: Element-level sampling---the selected pages are parsed by layout detection models into text, formula, and table blocks; each element type is independently clustered and assessed by CMCV, then sampled to balance both diversity and difficulty at the element granularity. The two levels are combined to produce the final training data for layout, text, formula, and table subtasks.}
\label{fig:ddas}
\end{figure*}

We propose Diversity-and-Difficulty-Aware Sampling (DDAS), which jointly optimizes diversity and difficulty at both page and element granularity. Central to DDAS is Cross-Model Consistency Verification (CMCV, detailed in \Cref{subsec:cmcv}), which leverages prediction agreement among heterogeneous models to classify samples into Easy/Medium/Hard difficulty tiers. The overall pipeline is shown in \Cref{fig:ddas}.

\paragraph{Stage 1: Page-level sampling.}
Pages in the document pool are embedded using ViT-base features (512-dim) and grouped via K-Means clustering. An initial uniform sample from each cluster is then evaluated by page-level CMCV (\Cref{subsec:cmcv}) to obtain difficulty labels. Based on the resulting difficulty distribution within each cluster, sampling weights are adjusted: clusters dominated by Easy samples receive lower weight, clusters with diverse difficulty distributions receive higher weight, and clusters dominated by invalid content (non-target languages, blank pages, \etc) are filtered out. Using the adjusted weights, we expand sampling from the original document pool to obtain the full page-level candidate set with CMCV difficulty annotations.

\paragraph{Stage 2: Element-level sampling.}
From the page-level candidate set, we extract individual elements (text, formula, and table blocks) using \mineru and PaddleOCR-VL layout detection models. For each element type, visual features are extracted and clustered independently, while element-level CMCV assigns difficulty labels. At this stage, all four subtasks---layout, text, formula, and table---have annotations along both the diversity (clustering) and difficulty (CMCV) dimensions.

\paragraph{Final sampling.}
Balanced sampling is performed in the joint cluster-difficulty space across all four subtasks: along the diversity dimension, large clusters are downsampled and small clusters are upsampled to correct long-tail shift; along the difficulty dimension, Medium and Hard samples are upweighted to enhance training signal informativeness. The final output is an SFT training set that covers all subtasks and balances diversity with difficulty.

By coupling clustering with CMCV at both page and element granularity, DDAS enables sampling decisions to simultaneously account for data distribution and training value, maximizing training signal density while controlling total data volume.

\subsection{Cross-Model Consistency Verification}
\label{subsec:cmcv}

DDAS relies on difficulty labels to guide sampling weight allocation, and subsequent annotation refinement and expert annotation also require difficulty information to determine resource investment strategies. However, ground truth is unavailable for massive unlabeled data. IMIC in \mineru~\cite{mineru25} and UACS in PaddleOCR-VL-1.5~\cite{paddleocr_vl15} use output consistency from multiple inferences of a single model as a difficulty proxy. This paradigm captures only the epistemic uncertainty of a single model and cannot distinguish between model-specific blind spots and universally hard problems---the former can be directly rectified via cross-model consensus, while the latter necessitates additional quality refinement or even human intervention. This distinction is critical for annotation strategy selection.

We propose Cross-Model Consistency Verification (CMCV), which extends difficulty assessment from single-model introspection to multi-model cross-validation. The underlying premise is intuitive and empirically supported: when multiple heterogeneous models produce consistent outputs for a given sample, the result is highly likely to be correct; when all models diverge substantially, the sample is genuinely difficult and none of the models can parse it reliably. Based on this premise, we run three heterogeneous document parsing models (\mineru~\cite{mineru25}, PaddleOCR-VL~\cite{cui2025paddleocrvl}, Qwen3-VL-30B~\cite{yang2025qwen3technicalreport}) independently on the candidate data produced by DDAS, compute task-specific pairwise consistency metrics (text: edit distance; table: TEDS; formula: CDM), and classify each sample into three difficulty tiers based on consistency patterns. Since \mineru is the target model to be improved, we anchor the difficulty taxonomy on its performance relative to external models:

\begin{itemize}
    \item \textbf{Easy}: \mineru's output is highly consistent with at least one external model. Model consensus indicates the parsing result is reliable, and any model's output can serve directly as annotation.
    \item \textbf{Medium}: The two external models agree with each other, but \mineru differs significantly from both. The external consensus serves as a reliable pseudo-label.
    \item \textbf{Hard}: All three models' outputs exhibit significant pairwise disagreement, and no reliable annotation can be obtained through model consensus.
\end{itemize}

These three data categories play different roles in training. Easy data is abundant and reliably annotated, forming the backbone for foundational capability building, but the model has largely mastered these scenarios and their marginal training value is limited. Medium data has the highest training value---it precisely pinpoints \mineru's capability gaps relative to peers, while the successful parsing by external models proves these samples are learnable, and the external consensus directly provides reliable annotations without further correction. Hard data is critical for capability breakthroughs, but its annotations are unreliable and require subsequent Judge-and-Refine correction or expert annotation (\Cref{subsec:annotation_pipeline}) before safe use. The respective strengths and constraints of these three categories naturally motivate the annotation pipeline described next.

CMCV thus enables rapid difficulty assessment on massive unlabeled data without human annotation, making large-scale data expansion and iteration feasible. Since Medium data is scarce but most valuable, we prioritize its proportion during DDAS sampling. The optimal ratio among the three categories varies by subtask---formula and table recognition are more sensitive to Hard samples, while text recognition benefits more from Medium samples.

\subsection{Annotation Pipeline for Hard Case}
\label{subsec:annotation_pipeline}

CMCV provides reliable automatic annotations for Easy and Medium samples. However, Hard samples---data on which all models fail to reach consensus---would introduce annotation noise that degrades rather than improves model performance if directly used for training. Improving annotation quality for these critical samples without relying on large-scale human annotation is the core challenge in advancing the Data Engine from ``filtering'' to ``refinement.'' To address this, we design a two-stage pipeline: an automated Judge-and-Refine correction loop followed by targeted expert annotation for residual failures.

\paragraph{Judge-and-Refine Annotation Pipeline.}

A natural approach to improving Hard sample annotations is to introduce test-time compute through an iterative judge-then-correct mechanism that lets the model examine and refine its own parsing results. However, naive self-reflection exhibits a systematic bias to accept its own outputs: when asked to check its output, the model tends to affirm the result as correct and overlook existing errors. The root cause lies in the asymmetry of cross-modal mapping---models excel at generating structured sequences from document images but struggle to infer visual appearance from structured sequences. For complex structural mappings such as \LaTeX{} formulas and HTML tables, the model cannot accurately judge how an output sequence will render visually in implicit space, significantly impairing its ability to detect structural errors.

To break this bottleneck, we incorporate \emph{render-then-verify} into the iterative correction loop: we compile \LaTeX{} formulas and render HTML tables into images, then feed both the original document image and the rendered image to the model as paired inputs alongside the judge-and-refine prompt. This design offers two advantages. First, it closes the missing mapping from structured text to visual layout, reducing the cross-modal reasoning burden. Second, the error-amplification effect of rendering translates subtle, text-domain structural flaws (\eg missing alignment symbols, unclosed tags) into salient visual anomalies or layout collapse, making defects readily detectable through visual comparison.

Based on this design, we build a visual-comparison-driven Judge-and-Refine iterative correction pipeline. The pipeline uses Qwen3-VL-235B as the Judge-Refine model---chosen for its strong multimodal reasoning capability and its independence from the CMCV model pool, which avoids systematic bias in error detection. Multi-round error localization and targeted correction proceed via direct visual comparison between the original document image and the rendered image. After processing through this pipeline, a subset of extremely complex cases still remains beyond automatic correction; these samples are routed to the expert annotation workflow.

\paragraph{Targeted Expert Annotation.}

For Hard samples that remain beyond automatic correction, we introduce expert human annotation to guarantee final quality. Annotation budget is allocated along two priority axes based on intermediate outputs from Judge-and-Refine:
\begin{enumerate}
    \item \emph{Correction efficiency}: Samples where the Judge stage has localized errors with high confidence but the Refine stage has failed to correct them receive top priority---annotators need only perform local corrections at identified locations, maximizing annotation throughput.
    \item \emph{Marginal impact}: Within the above pool, priority is further given to the subtask categories where the current model is weakest (determined by CMCV disagreement patterns), maximizing the marginal contribution of limited annotation budget to overall performance.
\end{enumerate}

Human annotation follows an AI pre-annotation and expert review-and-correction workflow. For pre-annotation, we use Gemini~3~Pro---chosen for its strong multimodal reasoning capability and its independence from the CMCV model pool, thereby avoiding data leakage. Automated QA tools further ensure annotation consistency. Compared to \mineru's human annotation process~\cite{mineru25}, annotation targets shift from random sampling to a precisely targeted subset identified through three-stage filtering, significantly improving annotation resource utilization.

The Data Engine produces a stratified dataset: approximately 65.5M Easy and Medium samples, automatically annotated via CMCV, are used for Stage~1 pre-training; 192K expert-annotated Hard samples are used for Stage~2 fine-tuning and Stage~3 GRPO alignment.

\section{Progressive Training Strategy}
\label{sec:training}

\minerupro inherits \mineru's~\cite{mineru25} 1.2B-parameter decoupled coarse-to-fine architecture (NaViT-675M vision encoder + Qwen2-0.5B language model) without any structural modification. The model is initialized from \mineru's Stage~0 checkpoint, which provides foundational vision-language alignment and OCR capabilities~\cite{mineru25}.

From this shared starting point, \minerupro employs a three-stage progressive training strategy that sequentially leverages data at different quality tiers produced by the Data Engine: Stage~1 pre-trains on large-scale CMCV auto-annotated data to build comprehensive foundational capabilities; Stage~2 fine-tunes on high-quality expert-annotated data to strengthen performance on hard scenarios; Stage~3 aligns output format and structural conventions through reinforcement learning. The three stages progress from data scale to data quality, with training configurations summarized in \Cref{tab:training_recipe}.

\begin{table}[t]
\centering
\caption{Training configurations for the three-stage progressive strategy. All three stages share the same model architecture and resolution settings; they differ in data source, data scale, and optimization hyperparameters, reflecting the progression from broad coverage (Stage~1) to targeted refinement (Stage~2) to metric-level alignment (Stage~3).}
\label{tab:training_recipe}
\small
\begin{tabular}{llccc}
\toprule
\textbf{Category} & \textbf{Parameter} & \textbf{Stage 1} & \textbf{Stage 2} & \textbf{Stage 3} \\
\midrule
\multirow{2}{*}{Vision} & Max Resolution & 2048$\times$28$\times$28 & 2048$\times$28$\times$28 & 2048$\times$28$\times$28 \\
 & \#Tokens per Image & 64--2048 & 64--2048 & 64--2048 \\
\midrule
\multirow{2}{*}{Data} & Dataset Type & \makecell{Layout \& OCR$^\dagger$ \\ \& Image Analysis} & \makecell{Layout \& OCR$^\dagger$ \\ \& Image Analysis} & \makecell{Layout \& Text \& \\ Formula \& Table} \\
 & \#Samples & 65.5M & \makecell{3.9M \\ (192K human-labeled)} & 192K \\
\midrule
\multirow{2}{*}{Model} & Trainable & All & All & All \\
 & Sequence Length & 8192 & 8192 & 8192 \\
\midrule
\multirow{4}{*}{Training} & Batch Size & 256 & 128 & 512 \\
 & ViT Learning Rate & $1 \times 10^{-4}$ & $5 \times 10^{-6}$ & $1 \times 10^{-7}$ \\
 & MLP/LLM Learning Rate & $1 \times 10^{-3}$ & $5 \times 10^{-5}$ & $1 \times 10^{-5}$ \\
 & Epoch & 1 & 1 & 1 \\
\bottomrule
\end{tabular}
\vspace{2pt}
\raggedright\footnotesize{$^\dagger$OCR collectively refers to text recognition, formula recognition, and table recognition.}
\end{table}

\subsection{Stage 1: Document Parsing Pre-training}
\label{subsec:stage1}

\paragraph{Training data.}
The training set consists of Easy and Medium samples produced by the Data Engine, with annotations derived from CMCV multi-model consensus. The data covers four core subtasks totaling approximately 65.5M samples: text recognition (21M), layout analysis (14M), formula recognition (13M), and table recognition (11.5M), plus 6M image analysis samples (charts, text-embedded images, \etc). Subtask ratios are adjusted based on their weights in the OmniDocBench overall score and the baseline model's per-task performance gaps.

\paragraph{Training configuration.}
All parameters are trainable. The language model uses a learning rate of $1 \times 10^{-3}$ and the vision encoder uses $1 \times 10^{-4}$, with a batch size of 256, and training runs for 1 epoch. Compared to \mineru's Stage~1 pre-training (6.9M samples/epoch $\times$ 2 epochs)~\cite{mineru25}, this stage expands data scale by nearly an order of magnitude (6.9M $\to$ 65.5M), with data quality also systematically improved through DDAS distribution correction and CMCV annotation filtering.

\subsection{Stage 2: High-Quality Supervised Fine-Tuning}
\label{subsec:stage2}

Stage~1 builds comprehensive foundational capabilities, but performance gaps persist on Hard samples. This stage uses high-quality expert-annotated data for targeted fine-tuning, strengthening hard scenarios while maintaining generalization on regular scenarios through mixed Stage~1 replay data.

\paragraph{Training data.}
The training set comprises two parts: (1) 192K high-quality Hard samples produced through the expert annotation pipeline; (2) replay data sampled proportionally from the Stage~1 training set to prevent catastrophic forgetting. The mixing ratio (Hard:Replay) is differentiated by subtask: layout analysis 6:1, text recognition 1:50, formula recognition 1:25, table recognition 1:10, and image analysis 1:4. This non-uniform mixing strategy reflects differences in hard sample volume and baseline performance across subtasks---layout analysis has more hard samples and a strong Stage~1 foundation, requiring less replay; text recognition has scarce hard samples and requires more replay to preserve generalization.

\paragraph{Training configuration.}
Building on the Stage~1 model, we adopt a lower learning rate of $5 \times 10^{-5}$ with a batch size of 128 for 1 epoch. The reduced learning rate protects foundational capabilities acquired in Stage~1 while fine-tuning decision boundaries on hard scenarios.

\subsection{Stage 3: Reinforcement Learning with GRPO}
\label{subsec:stage3}

The first two stages optimize content recognition accuracy through supervised learning. However, cross-entropy loss optimizes each token prediction independently and weights all tokens equally, without directly reflecting sequence-level or structure-level evaluation metrics (edit distance, CDM, TEDS, IoU). This stage bridges the gap between training objectives and evaluation metrics through reinforcement learning that directly optimizes task-level metrics.

We adopt Group Relative Policy Optimization (GRPO)~\cite{shao2024grpo} for alignment. For each input, $G$ groups of candidate outputs are sampled, rewards are computed directly using task-specific automatic evaluation metrics, and policy updates are guided by within-group relative advantages, eliminating the need for a separate reward model.

\paragraph{Reward design.}
Reward functions are designed separately for four subtasks, directly adopting the same metrics used in evaluation as reward signals: edit distance for text recognition, CDM for formula recognition, TEDS for table recognition, and category IoU for layout detection. This design directly aligns training optimization objectives with final evaluation metrics.

\paragraph{Training data.}
Training data is generated from Stage~2 model rollouts and filtered based on reward distribution: samples with excessively high rewards (model saturated, no effective learning signal) and excessively low rewards (samples too hard or annotations erroneous) are removed, retaining the mid-reward range to maximize effective policy gradient signal. All training data comes from the high-quality expert-annotated set to ensure reward signal reliability.

\paragraph{Training configuration.}
Building on the Stage~2 model, we adopt a learning rate of $1 \times 10^{-5}$ with a batch size of 512 for 1 epoch, sampling $G=16$ rollouts per sample. Following DAPO~\cite{yu2025dapo}, we apply clip-higher to stabilize advantage estimation and dynamic sampling to discard zero-variance rollout groups.

\section{OmniDocBench v1.6}
\label{sec:omnidocbench}

\subsection{Motivation}
\label{subsec:odb_motivation}

As leading document parsing models converge on OmniDocBench~v1.5, two fundamental issues limit its effectiveness as a benchmark:

\begin{figure*}[t]
\centering
\includegraphics[width=\textwidth]{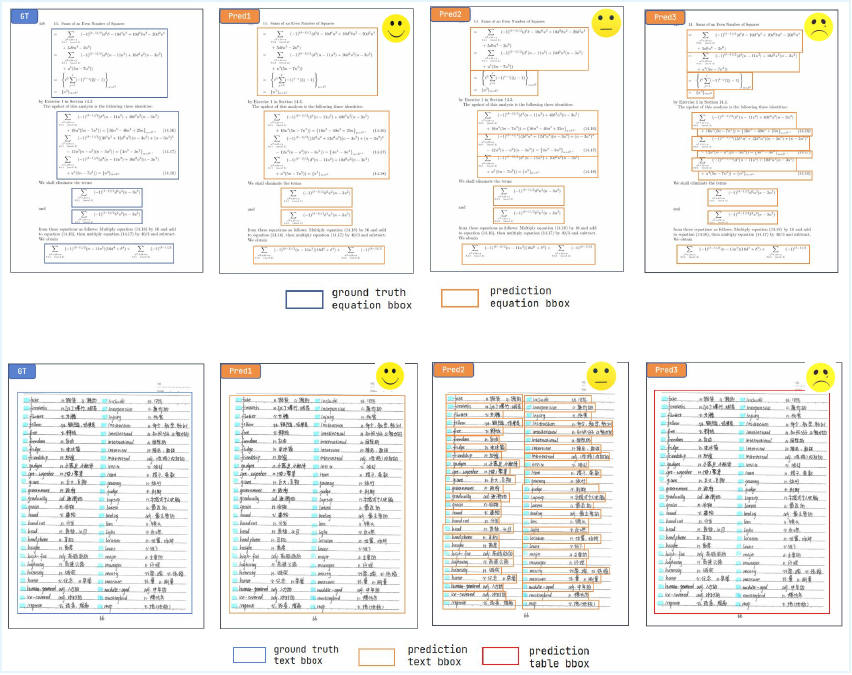}
\caption{Examples of element-matching bias in OmniDocBench~v1.5. Semantically correct predictions receive low scores due to granularity mismatch between predicted and ground-truth segmentation.}
\label{fig:mgam_example}
\end{figure*}

\paragraph{Matching strategy bias.}
v1.5 employs fixed-granularity one-to-one element matching, which silently penalizes systems whose output segmentation differs from the ground truth---even when the parsed content is entirely correct. As illustrated in \Cref{fig:mgam_example}, consider a multi-line formula annotated as a single block spanning $k$ lines: if a model produces the identical \LaTeX{} but segments it into $k{-}1$ or $k$ separate blocks, scores drop sharply from full marks to near zero despite semantically perfect output. A similar issue affects dense text: a region annotated as one block may be predicted line-by-line or even recognized as a table; in the latter case v1.5 assigns zero credit because no text element remains to match. These granularity-dependent scoring artifacts make cross-system comparisons unreliable.

\paragraph{Insufficient hard sample coverage.}
Through the large-scale difficulty stratification provided by our Data Engine (\Cref{sec:data_engine}), we find that samples labeled as Hard are virtually absent from the v1.5 evaluation set. The benchmark predominantly measures performance on low-to-medium difficulty documents, causing top models to cluster tightly with diminishing discriminative power.

To address these issues, we upgrade OmniDocBench to v1.6: we propose Multi-Granularity Adaptive Matching (MGAM) to eliminate matching bias (\Cref{subsec:mgam}), and expand the evaluation set with a dedicated Hard subset (\Cref{subsec:hard_subset_protocol}).

\subsection{Multi-Granularity Adaptive Matching}
\label{subsec:mgam}

We propose \textbf{Multi-Granularity Adaptive Matching (MGAM)}, which eliminates matching bias through adaptive granularity adjustment on the prediction side. The core principle is to keep the ground truth unchanged and search for the optimal segmentation granularity only on the prediction side. Given a set of ground truth elements $\mathcal{G} = \{g_1, g_2, \dots, g_m\}$ and prediction elements $\mathcal{P} = \{p_1, p_2, \dots, p_n\}$, MGAM generates candidate matching solutions through three stages and selects the global optimum:

\paragraph{Stage 1: Direct Bipartite Matching.}
We directly solve optimal bipartite matching between $\mathcal{P}$ and $\mathcal{G}$ at original granularity. Using the cost matrix $C_{ij} = 1 - \text{sim}(p_i, g_j)$ as input, the Hungarian algorithm solves for the minimum-cost matching $\mathcal{M}_1^* = \arg\min_{\mathcal{M}} \sum_{(i,j) \in \mathcal{M}} C_{ij}$, yielding the first candidate matching and its aggregate score $S_1$.

\paragraph{Stage 2: Prediction Splitting + Bipartite Matching.}
Each prediction element $p_i$ is split at \LaTeX{} line-break delimiters (\eg \texttt{\textbackslash\textbackslash}, \texttt{\textbackslash newline}, and equivalent symbols) to produce a fine-grained prediction set $\mathcal{P}' = \{p'_1, p'_2, \dots, p'_{n'}\}$ ($n' \geq n$). Prediction elements without splittable delimiters remain unchanged. Bipartite matching is re-solved on $\mathcal{P}'$ and $\mathcal{G}$, yielding candidate matching $\mathcal{M}_2^*$ and aggregate score $S_2$.

\paragraph{Stage 3: Partition Enumeration + Bipartite Matching.}
Stage~2 splitting may be too fine---annotation granularity is not necessarily per-line but may be any intermediate granularity between 1 and $k$ lines. To cover all possible merging schemes, we enumerate all valid ordered partitions of consecutive subsequences of $\mathcal{P}'$. Specifically, for $n'$ fine-grained prediction elements, there are $n'-1$ gaps between adjacent elements, and each gap can be either ``split'' or ``merged,'' producing $2^{n'-1}$ partition schemes. Each partition $\pi = (B_1, B_2, \dots, B_K)$ divides $\mathcal{P}'$ into $K$ consecutive blocks, where the $k$-th block is
\begin{equation}
B_k = \bigoplus_{t=l_k}^{r_k} p'_t
\end{equation}
with $\bigoplus$ denoting concatenation in original order. For each partition, we perform bipartite matching between the merged block set $\{B_1, \dots, B_K\}$ and $\mathcal{G}$, selecting the partition with the best matching score as candidate matching $\mathcal{M}_3^*$ and aggregate score $S_3$.

\paragraph{Global Optimum Selection.}
The final matching is the one with the best aggregate score among the three stages: $\mathcal{M}^* = \arg\max_{k \in \{1,2,3\}} S_k$, and the task-specific metric (\eg CDM for formulas, edit distance for text) is computed based on $\mathcal{M}^*$.

\paragraph{Dense text matching.}
The granularity mismatch problem is not limited to formulas. For dense text regions, prediction and annotation sides similarly differ in whether multiple text segments are merged into one large text box or split into multiple small ones. We reuse the MGAM algorithm for text elements with edit distance as the similarity metric. Additionally, if a model recognizes text in a region as a table (not uncommon for dense structured text), we convert the table back to plain text and include it in the same matching pipeline, avoiding unfair penalties due to format preference differences.

With MGAM, the evaluation becomes neutral to output granularity and format preferences, removing a systematic source of scoring variance across systems.

\subsection{Hard Subset and Three-Tier Evaluation Protocol}
\label{subsec:hard_subset_protocol}

To fill the coverage gap for hard scenarios, we construct a Hard subset of 296 pages selected from the pool of data labeled as Hard during Data Engine difficulty stratification. Samples are chosen to cover the most challenging scenario categories in document parsing, including complex nested tables, dense mathematical formula layouts, and unconventional layout structures. All Hard subset samples are excluded from every training stage of \minerupro (including Judge-and-Refine training data) and are annotated by professional teams with inter-annotator cross-validation to ensure ground truth quality.

OmniDocBench~v1.6 establishes a Base/Hard/Full three-tier evaluation protocol:
\begin{itemize}
    \item \textbf{Base} (1{,}355 pages): retains the original v1.5 evaluation set to maintain comparability with historical results.
    \item \textbf{Hard} (296 pages): the newly added hard sample subset, providing more sensitive measurement where top models saturate on standard evaluations.
    \item \textbf{Full} (1{,}651 pages): the complete union of both, providing comprehensive performance assessment.
\end{itemize}

\section{Experiments}
\label{sec:experiments}

This section evaluates \minerupro against both leading general-purpose VLMs and current SOTA document-parsing-specific models~\cite{gemini25pro,yang2025qwen3technicalreport,qwen35blog,glm_ocr,paddleocr_vl15,hunyuanocr,wu2026fireredocrtechnicalreport,deepseek_ocr,mineru25}. All competing models are re-evaluated under a unified environment using the same evaluation code.

\subsection{Evaluation Setup}
\label{subsec:eval_setup}

\paragraph{OmniDocBench v1.6.}
We evaluate on OmniDocBench~v1.6 using the Base/Hard/Full three-tier protocol described in \Cref{subsec:hard_subset_protocol}. The overall score follows the same formula as \mineru~\cite{mineru25}, averaging text (edit distance), table (TEDS), and formula (CDM) metrics. We additionally report sub-metrics: Text Edit$\downarrow$, Formula CDM$\uparrow$, Table TEDS$\uparrow$, Table TEDS-S$\uparrow$, Read Order Edit$\downarrow$.

\paragraph{Element-specific evaluation.}
To more accurately measure content recognition capability (excluding the confounding effect of layout detection errors), we crop document images based on ground truth layout boxes and separately evaluate text recognition, formula recognition, and table recognition as individual modules.

\subsection{End-to-End Document Parsing}
\label{subsec:e2e_results}

\begin{table*}[!t]
  \centering
  \caption{Performance comparison of document parsing methods on OmniDocBench~v1.6 Full across text, formula, table, and reading order extraction tasks.}
  \resizebox{1\textwidth}{!}{
    \begin{tabular}{c|l l|c|c c c c c}
    \toprule
    \textbf{Model Type} & \textbf{Methods} & \textbf{Param} & \textbf{Overall}$\uparrow$ & \textbf{Text\textsuperscript{Edit}}$\downarrow$ & \textbf{Formula\textsuperscript{CDM}}$\uparrow$ & \textbf{Table\textsuperscript{TEDS}}$\uparrow$ & \textbf{Table\textsuperscript{TEDS-S}}$\uparrow$ & \textbf{Read Order\textsuperscript{Edit}}$\downarrow$ \\
    \midrule
    \multirow{15}{*}{\makecell{\textbf{Specialized}\\\textbf{VLMs}}} 
    & \cellcolor{gray!20}\textbf{\minerupro} & \cellcolor{gray!20}1.2B & \cellcolor{gray!20}\textbf{95.69} & \cellcolor{gray!20}\underline{0.036} & \cellcolor{gray!20}\textbf{97.29} & \cellcolor{gray!20}\textbf{93.42} & \cellcolor{gray!20}\textbf{95.92} & \cellcolor{gray!20}\underline{0.120} \\
    & GLM-OCR~\cite{glm_ocr} & 0.9B & \underline{95.15} & 0.044 & \underline{96.99} & \underline{92.83} & \underline{95.39} & 0.133 \\
    & PaddleOCR-VL-1.5~\cite{paddleocr_vl15} & 0.9B & 94.87 & 0.038 & 96.69 & 91.67 & 94.37 & 0.130 \\
    & PaddleOCR-VL~\cite{cui2025paddleocrvl} & 0.9B & 94.11 & 0.040 & 95.70 & 90.65 & 93.74 & 0.135 \\
    & Youtu-Parsing~\cite{yin2026youtuparsingperceptionstructuringrecognition} & 2.5B & 93.68 & 0.044 & 93.45 & 92.02 & 95.00 & \textbf{0.116} \\
    & Logics-Parsing-v2~\cite{chen2025logicsparsingtechnicalreport} & 4B & 93.27 & 0.041 & 95.47 & 88.42 & 91.98 & 0.137 \\
    & FireRed-OCR~\cite{wu2026fireredocrtechnicalreport} & 2B & 93.20 & 0.037 & 95.27 & 88.04 & 91.06 & 0.131 \\
    & \mineru~\cite{mineru25} & 1.2B & 92.98 & 0.045 & 95.59 & 87.88 & 91.47 & 0.130 \\
    & OpenDoc-0.1B~\cite{du2025unirec} & 0.1B & 90.64 & 0.049 & 92.93 & 83.88 & 87.45 & 0.140 \\
    & dots.ocr~\cite{dots_ocr} & 3B & 90.50 & 0.048 & 89.12 & 87.18 & 90.58 & 0.138 \\
    & DeepSeek-OCR 2~\cite{deepseek_ocr} & 3B & 90.17 & 0.050 & 91.59 & 83.89 & 87.75 & 0.144 \\
    & HunyuanOCR~\cite{hunyuanocr} & 1B & 89.87 & 0.089 & 87.44 & 91.01 & 93.23 & 0.171 \\  
    & Dolphin-v2~\cite{dolphin} & 3B & 89.34 & 0.069 & 90.53 & 84.40 & 87.44 & 0.150 \\
    & OCRVerse~\cite{zhong2026ocrverseholisticocrendtoend} & 4B & 88.44 & 0.063 & 89.14 & 82.44 & 86.27 & 0.163 \\
    & MonkeyOCR-pro-3B~\cite{monkeyocr} & 3B & 88.43 & 0.074 & 88.33 & 84.35 & 88.62 & 0.189 \\
    \midrule
    \multirow{6}{*}{\makecell{\textbf{General}\\\textbf{VLMs}}}
    & Ovis2.6-30B-A3B~\cite{lu2024ovisstructuralembeddingalignment,lu2025ovis25technicalreport} & 30B & 93.62 & \textbf{0.035} & 94.93 & 89.44 & 92.40 & 0.135 \\
    & Gemini 3 Pro & -- & 92.85 & 0.064 & 95.83 & 89.15 & 92.96 & 0.165 \\
    & Gemini 3 Flash & -- & 92.58 & 0.066 & 95.03 & 89.29 & 93.51 & 0.173 \\
    & Qwen3-VL-235B~\cite{yang2025qwen3technicalreport} & 235B & 89.78 & 0.063 & 92.53 & 83.07 & 86.75 & 0.166 \\
    & GPT-5.2 & -- & 86.52 & 0.114 & 88.00 & 82.95 & 87.93 & 0.193 \\
    & InternVL3.5-241B~\cite{wang2025internvl35advancingopensourcemultimodal} & 241B & 83.61 & 0.130 & 89.52 & 74.35 & 79.78 & 0.215 \\
    \bottomrule
    \end{tabular}%
  }
  \label{tab:main_results}
  \vspace{-4pt}
\end{table*}

As shown in \Cref{tab:main_results}, \minerupro ranks first on Full with 95.69, improving over the same-architecture \mineru baseline (92.98) by 2.71 points---confirming that all gains are data-driven. On the Base subset~\Cref{tab:base_detailed}, the top three models (GLM-OCR 96.19, \minerupro 96.12, PaddleOCR-VL-1.5 95.72) are within 0.5 points, indicating near-saturation on standard scenarios. On the Hard subset~\Cref{tab:hard_detailed}, \minerupro leads at 94.08, exceeding both GLM-OCR and PaddleOCR-VL-1.5 (both at 92.01) by 2.07 points, demonstrating the Data Engine's advantage in hard scenario robustness and validating the Hard subset's discriminative power.

Across sub-metrics, \minerupro achieves the best scores in formula recognition (CDM 97.29), table recognition (TEDS 93.42, TEDS-S 95.92), and reading order (0.120). Notably, Gemini 3 Pro/Flash benefit substantially from the corrected matching in OmniDocBench~v1.6 (Full 92.85/92.58), narrowing the gap with specialized models, yet specialized models at only 0.9B--1.2B parameters maintain an overall lead.

\textbf{Training stage ablation.}
\Cref{tab:stage_ablation} reports the incremental contribution of each training stage.

\begin{table}[t]
\centering
\caption{Training stage ablation on OmniDocBench~v1.6.}
\label{tab:stage_ablation}
\small
\setlength{\tabcolsep}{3pt}
\begin{tabular}{lcccccccc}
\toprule
\textbf{Stage} & \textbf{Base} & \textbf{Hard} & \textbf{Full} & $\Delta$\textbf{Full} & \textbf{Text$\downarrow$} & \textbf{CDM$\uparrow$} & \textbf{TEDS$\uparrow$} \\
\midrule
\mineru (baseline) & 93.23 & 91.65 & 92.98 & --- & 0.045 & 95.59 & 87.88 \\
Stage 1: Large-Scale SFT & 94.54 & 93.10 & 94.29 & +1.31 & 0.039 & 96.40 & 90.37 \\
+ Stage 2: Hard-Sample SFT & 95.60 & 93.84 & 95.25 & +0.96 & 0.036 & 96.48 & 92.87 \\
+ Stage 3: GRPO & \textbf{96.12} & \textbf{94.08} & \textbf{95.69} & +0.45 & \textbf{0.036} & \textbf{97.29} & \textbf{93.42} \\
\bottomrule
\end{tabular}
\end{table}

Stage~1 (large-scale SFT) contributes the largest single-stage gain (+1.31), indicating that the Data Engine's optimization of data coverage and annotation quality is the primary driver of performance improvement. Stage~2 (hard sample fine-tuning) adds +0.96, with the most notable contribution in table recognition (TEDS 90.37 $\to$ 92.87, +2.50). Stage~3 (GRPO) contributes +0.45, primarily reflected in formula CDM improvement (96.48 $\to$ 97.29, +0.81), driven by reinforcement learning's direct optimization of task-level metrics. The cumulative improvement on the Hard subset (91.65 $\to$ 94.08, +2.43) is comparable to the Base subset (93.23 $\to$ 96.12, +2.89), indicating that the progressive training strategy achieves balanced capability improvement across both hard and standard scenarios.

\subsection{Element-Specific Parsing}
\label{subsec:element_results}

Layout detection accuracy in end-to-end evaluation cascades into content recognition scores, and differences in output granularity and segmentation strategies across models prevent precise matching of a small number of elements. To more fairly evaluate pure content recognition capability, we crop document images based on ground truth layout boxes and test text, formula, and table recognition as individual modules. Note that end-to-end models do not receive element category priors in this setting, which may partially explain their larger performance gap compared to decoupled two-stage models.

\textbf{Text recognition.}
As shown in \Cref{tab:text_results}, \minerupro ranks first on Full with an edit distance of 0.019, a 30.5\% reduction from the \mineru baseline (0.028). Hundred-billion-scale general VLMs (Qwen3.5-397B, Qwen3-VL-235B) demonstrate competitive text recognition performance comparable to specialized models, while end-to-end models (DeepSeek-OCR~2, FireRed-OCR) show significant degradation without category priors.

\begin{table}[t]
\centering
\caption{Text recognition (Edit Distance$\downarrow$) on OmniDocBench~v1.6.}
\label{tab:text_results}
\small
\begin{tabular}{llccc}
\toprule
\textbf{Model} & \textbf{Type} & \textbf{Base$\downarrow$} & \textbf{Hard$\downarrow$} & \textbf{Full$\downarrow$} \\
\midrule
\textbf{\minerupro} & Decoupled & \textbf{0.015} & \textbf{0.048} & \textbf{0.019} \\
Qwen3.5-397B~\cite{qwen35blog} & General & \underline{0.016} & 0.052 & \underline{0.020} \\
GLM-OCR~\cite{glm_ocr} & Decoupled & 0.016 & 0.053 & 0.021 \\
Qwen3-VL-235B~\cite{yang2025qwen3technicalreport} & General & 0.017 & \underline{0.049} & 0.021 \\
PaddleOCR-VL-1.5~\cite{paddleocr_vl15} & Decoupled & 0.018 & 0.056 & 0.022 \\
\mineru~\cite{mineru25} & Decoupled & 0.023 & 0.066 & 0.028 \\
PaddleOCR-VL~\cite{cui2025paddleocrvl} & Decoupled & 0.019 & 0.057 & 0.023 \\
DeepSeek-OCR 2~\cite{deepseek_ocr} & End-to-End & 0.057 & 0.130 & 0.066 \\
FireRed-OCR~\cite{wu2026fireredocrtechnicalreport} & End-to-End & 0.135 & 0.176 & 0.140 \\
\bottomrule
\end{tabular}
\end{table}

\begin{table*}[t]
\centering
\caption{Formula recognition (CDM$\uparrow$) across multiple benchmarks. CPE, HWE, SCE, and SPE are the Complex Printed, Handwritten, Screen-Captured, and Simple Printed Expression subsets from UniMERNet~\cite{wang2024unimernet}, respectively.}
\label{tab:formula_results}
\small
\resizebox{0.9 \textwidth}{!}{
\setlength{\tabcolsep}{4pt}
\begin{tabular}{lccccccccc}
\toprule
\multirow{2}{*}{\textbf{Model}} & \multicolumn{2}{c}{\textbf{OmniDoc}} & \multicolumn{5}{c}{\textbf{Public}} & \multicolumn{2}{c}{\textbf{Inhouse}} \\ \cmidrule(lr){2-3} \cmidrule(lr){4-8} \cmidrule(lr){9-10} &  \textbf{Base} & \textbf{Hard} & \textbf{CPE} & \textbf{HWE} & \textbf{SCE} & \textbf{SPE} & \textbf{LaTeX-80M} & \textbf{Chinese} & \textbf{Fuzzy} \\
\midrule
\textbf{\minerupro} & \textbf{99.20} & \textbf{98.79} & \textbf{98.97} & \underline{95.38} & \underline{97.04} & \textbf{99.44} & \textbf{97.23} & \underline{95.28} & \underline{94.90} \\
PaddleOCR-VL-1.5~\cite{paddleocr_vl15} & \underline{98.76} & 97.22 & 98.84 & 92.27 & 94.95 & 99.27 & 92.77 & 94.06 & 89.73 \\
GLM-OCR~\cite{glm_ocr} & 98.75 & 98.28 & 96.74 & 95.10 & \textbf{97.77} & 98.42 & 95.39 & 94.35 & 93.75 \\
PaddleOCR-VL~\cite{cui2025paddleocrvl} & 98.72 & 97.64 & \underline{98.93} & 94.45 & 95.88 & 99.30 & 93.67 & 94.35 & 91.56 \\
Qwen3.5-397B~\cite{qwen35blog} & 98.19 & 97.25 & 98.32 & \textbf{97.59} & 95.87 & \underline{99.41} & 95.17 & 78.24 & 90.53 \\
Qwen3-VL-235B~\cite{yang2025qwen3technicalreport} & 97.72 & 98.13 & 97.47 & 94.23 & 96.21 & 98.46 & 95.33 & 92.69 & 93.59 \\
\mineru~\cite{mineru25} & 97.25 & \underline{98.67} & 97.79 & 94.42 & 96.65 & 98.57 & \underline{96.23} & \textbf{95.50} & \textbf{94.92} \\
FireRed-OCR~\cite{wu2026fireredocrtechnicalreport} & 96.71 & 94.54 & 94.35 & 85.42 & 89.94 & 96.75 & 83.41 & 87.94 & 87.77 \\
DeepSeek-OCR 2~\cite{deepseek_ocr} & 95.95 & 93.39 & 91.97 & 81.67 & 77.19 & 95.51 & 72.04 & 87.82 & 85.13 \\
\bottomrule
\end{tabular}
}
\end{table*}

\textbf{Formula recognition.}
\Cref{tab:formula_results} reports CDM scores across 9 benchmarks. \minerupro achieves the best score in 5 dimensions and the second-highest in the remaining four. Specifically, it falls short on HWE (handwritten formulas) against Qwen3.5-397B (95.38 \vs 97.59) and on SCE against GLM-OCR (97.04 \vs 97.77), while only trailing slightly behind \mineru on Chinese and Fuzzy subsets. On OmniDocBench Base, CDM reaches 99.20 (out of 100), approaching the performance ceiling for formula recognition. While Qwen3.5-397B excels on handwritten formulas, it reveals a notable weakness on Chinese formulas (Chinese 78.24).

\textbf{Table recognition.}
As shown in \Cref{tab:table_results}, \minerupro ranks first in both Overall TEDS (91.10) and TEDS-S (94.48), improving over \mineru by 3.16 and 2.31 percentage points, respectively. The advantage is most pronounced on the Hard subset (TEDS 92.46 \vs \mineru's 88.28, +4.18), indicating that the Data Engine's hard sample mining and expert annotation contribute most to table recognition. GLM-OCR is slightly better on OmniDocBench Base (96.14) and CCOCR (89.17) but is less stable than \minerupro across benchmarks. PaddleOCR-VL-1.5 shows notable performance drops on CCOCR (TEDS 76.34) and Inhouse (TEDS 72.66), suggesting limited table recognition generalization.

\begin{table*}[t]
\centering
\caption{Table recognition (TEDS \& TEDS-S$\uparrow$) across multiple benchmarks.}
\label{tab:table_results}
\small
\resizebox{0.97 \textwidth}{!}{
\setlength{\tabcolsep}{3pt}
\begin{tabular}{llcccccccccccc}
\toprule
\multirow{2}{*}{\textbf{Model}} & \multirow{2}{*}{\textbf{Type}} & \multicolumn{2}{c}{\textbf{OmniDoc Base}} & \multicolumn{2}{c}{\textbf{OmniDoc Hard}} & \multicolumn{2}{c}{\textbf{CCOCR}} & \multicolumn{2}{c}{\textbf{OCRBv2}} & \multicolumn{2}{c}{\textbf{Inhouse}} & \multicolumn{2}{c}{\textbf{Overall}} \\
\cmidrule(lr){3-4} \cmidrule(lr){5-6} \cmidrule(lr){7-8} \cmidrule(lr){9-10} \cmidrule(lr){11-12} \cmidrule(lr){13-14}
 & & TEDS & TEDS-S & TEDS & TEDS-S & TEDS & TEDS-S & TEDS & TEDS-S & TEDS & TEDS-S & TEDS & TEDS-S \\
\midrule
\textbf{\minerupro} & Decoup. & \underline{95.67} & \underline{97.42} & \textbf{92.46} & \textbf{94.67} & 88.49 & \underline{91.90} & \textbf{93.56} & \textbf{96.27} & \textbf{82.70} & \textbf{89.65} & \textbf{91.10} & \textbf{94.48} \\
GLM-OCR~\cite{glm_ocr} & Decoup. & \textbf{96.14} & \textbf{97.60} & \underline{90.49} & \underline{93.47} & \textbf{89.17} & \textbf{92.58} & 91.19 & 94.44 & \underline{79.41} & \underline{87.65} & \underline{89.71} & \underline{93.52} \\
Gemini 3 Pro & General & 94.42 & 97.37 & 88.16 & 91.34 & 86.47 & 90.10 & 91.73 & 94.96 & 75.91 & 85.26 & 88.21 & 92.65 \\
PaddleOCR-VL-1.5~\cite{paddleocr_vl15} & Decoup. & 93.85 & 95.57 & 88.28 & 91.79 & 76.34 & 81.38 & 82.64 & 86.91 & 72.66 & 81.93 & 82.91 & 87.60 \\
Qwen3.5-397B~\cite{qwen35blog} & General & 93.76 & 96.27 & 89.67 & 93.19 & \underline{88.87} & 91.89 & 88.11 & 91.21 & 77.49 & 85.53 & 87.60 & 91.57 \\
Qwen3-VL-235B~\cite{yang2025qwen3technicalreport} & General & 92.92 & 95.48 & 87.55 & 91.80 & 87.38 & 91.22 & 88.81 & 92.94 & 74.39 & 83.94 & 86.64 & 91.44 \\
\mineru~\cite{mineru25} & Decoup. & 92.87 & 95.33 & 88.28 & 91.80 & 84.35 & 88.25 & \underline{92.32} & \underline{95.31} & 76.95 & 85.98 & 87.94 & 92.17 \\
PaddleOCR-VL~\cite{cui2025paddleocrvl} & Decoup. & 92.31 & 94.64 & 89.21 & 92.02 & 81.54 & 85.80 & 82.29 & 85.79 & 74.88 & 82.92 & 83.67 & 87.85 \\
FireRed-OCR~\cite{wu2026fireredocrtechnicalreport} & E2E & 88.12 & 90.29 & 86.11 & 89.58 & 82.24 & 86.62 & 86.93 & 90.34 & 69.57 & 79.02 & 83.02 & 87.47 \\
DeepSeek-OCR 2~\cite{deepseek_ocr} & E2E & 79.27 & 82.46 & 75.07 & 80.06 & 66.98 & 72.74 & 84.38 & 88.70 & 57.70 & 69.35 & 74.59 & 80.35 \\
\bottomrule
\end{tabular}
}
\end{table*}

\section{Conclusion}
\label{sec:conclusion}

We present \minerupro, which improves the OmniDocBench~v1.6 overall score from 92.98 to 95.69 solely through systematic data engineering while keeping the 1.2B-parameter model architecture completely fixed, surpassing all existing methods. This result demonstrates that at the current stage where architectures are maturing, co-optimizing training data coverage, informativeness, and annotation accuracy yields greater performance gains than architectural improvements alone. To this end, we contribute a Data Engine that expands training data from under 10M to 65.5M pages while systematically improving annotation quality, a three-stage progressive training strategy matched to data quality tiers, and the OmniDocBench~v1.6 three-tier evaluation protocol that corrects evaluation biases. These tools and methodologies provide the community with a performance improvement pathway that is orthogonal to and complementary with architectural innovation.

\subsection*{Limitations and Future Directions}

\paragraph{Fundamental challenges in evaluation.}
OmniDocBench~v1.6 improves scoring fairness through corrected matching strategies, but the element-matching paradigm itself has inherent limitations. The ambiguity is twofold: at the format level, the same content can be expressed in multiple equivalent notations (\eg HTML \vs Markdown for tables, different \LaTeX{} commands for the same formula); at the structural level, the same visual layout can be legitimately represented with different element types---for instance, a bilingual word list with aligned Chinese and English columns is equally valid as line-by-line text pairs or as a two-column table, and even human annotators may disagree on which representation is ``correct.'' Developing semantic-equivalence-aware evaluation methods that account for both format and structural ambiguity remains an open problem.

\paragraph{Evaluation coverage and domain adaptation.}
OmniDocBench~v1.6 aims to cover mainstream application scenarios; for vertical domains with higher precision requirements (\eg finance, legal, medical), constructing domain-specific evaluation sets is a necessary complement. Furthermore, as model capabilities approach human-level performance, ensuring the precision of evaluation set annotations themselves becomes an increasingly pressing challenge.

\paragraph{From parsing accuracy to structural understanding.}
This work focuses on content accuracy in document parsing. However, for downstream applications, structural relationships within documents---such as hierarchical relationships between headings and body text, semantic bindings between figures/tables and referring text, and cross-page content continuity---are equally critical for document retrieval and downstream semantic understanding. Advancing parsing from ``content extraction'' to ``structured semantic understanding'' represents a natural next step for document parsing research.

\clearpage

\clearpage
\newpage
\bibliographystyle{plainnat}
\setcitestyle{numbers}
\bibliography{paper}

@article{olmocr,
  title={olmOCR: Unlocking Trillions of Tokens in PDFs with Vision Language Models},
  author={Poznanski, Jake and Rangapur, Aman and Borchardt, Jon and Dunkelberger, Jason and Huff, Regan and Lin, Daniel and Wilhelm, Christopher and Lo, Kyle and Soldaini, Luca},
  journal={arXiv preprint arXiv:2502.18443},
  year={2025}
}

@article{mineru,
  title={MinerU: An Open-Source Solution for Precise Document Content Extraction},
  author={Wang, Bin and Xu, Chao and Zhao, Xiaomeng and Ouyang, Linke and Wu, Fan and Zhao, Zhiyuan and Xu, Rui and Liu, Kaiwen and Qu, Yuan and Shang, Fukai and Zhang, Bo and Wei, Liqun and Sui, Zhihao and Li, Wei and Shi, Botian and Qiao, Yu and Lin, Dahua and He, Conghui},
  journal={arXiv preprint arXiv:2409.18839},
  year={2024}
}

@article{rag_anything,
  title={RAG-Anything: All-in-One RAG Framework},
  author={Guo, Zirui and Ren, Xubin and Xu, Lingrui and Zhang, Jiahao and Huang, Chao},
  journal={arXiv preprint arXiv:2510.12323},
  year={2025}
}

@article{zhang2024document,
  title={Document parsing unveiled: Techniques, challenges, and prospects for structured information extraction},
  author={Zhang, Qintong and Wang, Bin and Huang, Victor Shea-Jay and Zhang, Junyuan and Wang, Zhengren and Liang, Hao and He, Conghui and Zhang, Wentao},
  journal={arXiv preprint arXiv:2410.21169},
  year={2024}
}

@inproceedings{zhang2025ocr,
  title={Ocr hinders rag: Evaluating the cascading impact of ocr on retrieval-augmented generation},
  author={Zhang, Junyuan and Zhang, Qintong and Wang, Bin and Ouyang, Linke and Wen, Zichen and Li, Ying and Chow, Ka-Ho and He, Conghui and Zhang, Wentao},
  booktitle={Proceedings of the IEEE/CVF International Conference on Computer Vision},
  pages={17443--17453},
  year={2025}
}

@article{bookrag,
  title={BookRAG: A Hierarchical Structure-aware Index-based Approach for Retrieval-Augmented Generation on Complex Documents},
  author={Wang, Shu and Zhou, Yingli and Fang, Yixiang},
  journal={arXiv preprint arXiv:2512.03413},
  year={2025}
}

@article{nougat,
  title={Nougat: Neural Optical Understanding for Academic Documents},
  author={Blecher, Lukas and Cucurull, Guillem and Scialom, Thomas and Stojnic, Robert},
  journal={arXiv preprint arXiv:2308.13418},
  year={2023}
}

@article{got_ocr,
  title={General OCR Theory: Towards OCR-2.0 via a Unified End-to-end Model},
  author={Wei, Haoran and Liu, Chenglong and Chen, Jinyue and Wang, Jia and Kong, Lingyu and Xu, Yanming and Ge, Zheng and Zhao, Liang and Sun, Jianjian and Peng, Yuang and Han, Chunrui and Zhang, Xiangyu},
  journal={arXiv preprint arXiv:2409.01704},
  year={2024}
}

@article{mineru25,
  title={MinerU2.5: A Decoupled Vision-Language Model for Efficient High-Resolution Document Parsing},
  author={Niu, Junbo and Liu, Zheng and Gu, Zhuangcheng and Wang, Bin and Ouyang, Linke and Zhao, Zhiyuan and Chu, Tao and He, Tianyao and Wu, Fan and Zhang, Qintong and Jin, Zhenjiang and Liang, Guang and Zhang, Rui and Zhang, Wenzheng and Qu, Yuan and Ren, Zhifei and Sun, Yuefeng and Zheng, Yuanhong and Ma, Dongsheng and Tang, Zirui and Niu, Boyu and Miao, Ziyang and Dong, Hejun and Qian, Siyi and Zhang, Junyuan and Chen, Jingzhou and Wang, Fangdong and Zhao, Xiaomeng and Wei, Liqun and Li, Wei and Wang, Shasha and Xu, Ruiliang and Cao, Yuanyuan and Chen, Lu and Wu, Qianqian and Gu, Huaiyu and Lu, Lindong and Wang, Keming and Lin, Dechen and Shen, Guanlin and Zhou, Xuanhe and Zhang, Linfeng and Zang, Yuhang and Dong, Xiaoyi and Wang, Jiaqi and Zhang, Bo and Bai, Lei and Chu, Pei and Li, Weijia and Wu, Jiang and Wu, Lijun and Li, Zhenxiang and Wang, Guangyu and Tu, Zhongying and Xu, Chao and Chen, Kai and Qiao, Yu and Zhou, Bowen and Lin, Dahua and Zhang, Wentao and He, Conghui},
  journal={arXiv preprint arXiv:2509.22186},
  year={2025}
}

@article{minerudiffusion,
  title={MinerU-Diffusion: Rethinking Document OCR as Inverse Rendering via Diffusion Decoding},
  author={Dong, Hejun and Niu, Junbo and Wang, Bin and Zeng, Weijun and Zhang, Wentao and He, Conghui},
  journal={arXiv preprint arXiv:2603.22458},
  year={2026}
}

@article{livathinos2025,
  title={Docling: An efficient open-source toolkit for ai-driven document conversion},
  author={Livathinos, Nikolaos and Auer, Christoph and Lysak, Maksym and Nassar, Ahmed and Dolfi, Michele and Vagenas, Panos and Ramis, Cesar Berrospi and Omenetti, Matteo and Dinkla, Kasper and Kim, Yusik and others},
  journal={arXiv preprint arXiv:2501.17887},
  year={2025}
}

@misc{marker,
  author       = {Vik Paruchuri},
  title        = {Marker},
  year         = {2025},
  howpublished = {\url{https://github.com/datalab-to/marker}},
  note         = {Accessed:2025-09-25},
}

@article{cui2025paddleocr,
  title={Paddleocr 3.0 technical report},
  author={Cui, Cheng and Sun, Ting and Lin, Manhui and Gao, Tingquan and Zhang, Yubo and Liu, Jiaxuan and Wang, Xueqing and Zhang, Zelun and Zhou, Changda and Liu, Hongen and others},
  journal={arXiv preprint arXiv:2507.05595},
  year={2025}
}

@article{cui2025paddleocrvl,
  title={Paddleocr-vl: Boosting multilingual document parsing via a 0.9 b ultra-compact vision-language model},
  author={Cui, Cheng and Sun, Ting and Liang, Suyin and Gao, Tingquan and Zhang, Zelun and Liu, Jiaxuan and Wang, Xueqing and Zhou, Changda and Liu, Hongen and Lin, Manhui and others},
  journal={arXiv preprint arXiv:2510.14528},
  year={2025}
}

@article{donut,
  title={Donut: Document understanding transformer without ocr},
  author={Kim, Geewook and Hong, Teakgyu and Yim, Moonbin and Park, Jinyoung and Yim, Jinyeong and Hwang, Wonseok and Yun, Sangdoo and Han, Dongyoon and Park, Seunghyun},
  journal={arXiv preprint arXiv:2111.15664},
  volume={7},
  number={15},
  pages={2},
  year={2021}
}

@article{ocean_ocr,
  title={Ocean-ocr: Towards general ocr application via a vision-language model},
  author={Chen, Song and Guo, Xinyu and Li, Yadong and Zhang, Tao and Lin, Mingan and Kuang, Dongdong and Zhang, Youwei and Ming, Lingfeng and Zhang, Fengyu and Wang, Yuran and others},
  journal={arXiv preprint arXiv:2501.15558},
  year={2025}
}

@article{dots_ocr,
  title={dots. ocr: Multilingual document layout parsing in a single vision-language model},
  author={Li, Yumeng and Yang, Guang and Liu, Hao and Wang, Bowen and Zhang, Colin},
  journal={arXiv preprint arXiv:2512.02498},
  year={2025}
}

@article{monkeyocr,
  title={MonkeyOCR: Document Parsing with a Structure-Recognition-Relation Triplet Paradigm},
  author={Li, Zhang and Liu, Yuliang and Liu, Qiang and Ma, Zhiyin and Zhang, Ziyang and Zhang, Shuo and Guo, Zidun and Zhang, Jiarui and Wang, Xinyu and Bai, Xiang},
  journal={arXiv preprint arXiv:2506.05218},
  year={2025}
}

@article{glm_ocr,
  title={Glm-ocr technical report},
  author={Duan, Shuaiqi and Xue, Yadong and Wang, Weihan and Su, Zhe and Liu, Huan and Yang, Sheng and Gan, Guobing and Wang, Guo and Wang, Zihan and Yan, Shengdong and others},
  journal={arXiv preprint arXiv:2603.10910},
  year={2026}
}

@article{paddleocr_vl15,
  title={PaddleOCR-VL-1.5: Towards a Multi-Task 0.9 B VLM for Robust In-the-Wild Document Parsing},
  author={Cui, Cheng and Sun, Ting and Liang, Suyin and Gao, Tingquan and Zhang, Zelun and Liu, Jiaxuan and Wang, Xueqing and Zhou, Changda and Liu, Hongen and Lin, Manhui and others},
  journal={arXiv preprint arXiv:2601.21957},
  year={2026}
}

@article{hunyuanocr,
  title={Hunyuanocr technical report},
  author={Team, Hunyuan Vision and Lyu, Pengyuan and Wan, Xingyu and Li, Gengluo and Peng, Shangpin and Wang, Weinong and Wu, Liang and Shen, Huawen and Zhou, Yu and Tang, Canhui and others},
  journal={arXiv preprint arXiv:2511.19575},
  year={2025}
}

@article{deepseek_ocr,
  title={Deepseek-ocr: Contexts optical compression},
  author={Wei, Haoran and Sun, Yaofeng and Li, Yukun},
  journal={arXiv preprint arXiv:2510.18234},
  year={2025}
}

@article{gemini25pro,
  title={Gemini 2.5: Pushing the frontier with advanced reasoning, multimodality, long context, and next generation agentic capabilities},
  author={Comanici, Gheorghe and Bieber, Eric and Schaekermann, Mike and Pasupat, Ice and Sachdeva, Noveen and Dhillon, Inderjit and Blistein, Marcel and Ram, Ori and Zhang, Dan and Rosen, Evan and others},
  journal={arXiv preprint arXiv:2507.06261},
  year={2025}
}

@article{qwen25vl,
  title={Qwen2. 5-vl technical report},
  author={Bai, Shuai and Chen, Keqin and Liu, Xuejing and Wang, Jialin and Ge, Wenbin and Song, Sibo and Dang, Kai and Wang, Peng and Wang, Shijie and Tang, Jun and others},
  journal={arXiv preprint arXiv:2502.13923},
  year={2025}
}

@article{nativeres,
  title={Native visual understanding: Resolving resolution dilemmas in vision-language models},
  author={Niu, Junbo and Zheng, Yuanhong and Miao, Ziyang and Dong, Hejun and Ge, Chunjiang and Liang, Hao and Lu, Ma and Zeng, Bohan and Zheng, Qiahao and He, Conghui and others},
  journal={arXiv preprint arXiv:2506.12776},
  year={2025}
}

@article{ng2021datacentric,
  title={Data-Centric AI Competition},
  author={Ng, Andrew and Laird, Dillon and He, Lynn},
  journal={DeepLearning AI. Available online: https://https-deeplearning-ai. github. io/data-centric-comp/(accessed on 9 December 2021)},
  year={2021}
}

@inproceedings{zha2023datacentric,
  title={Data-centric ai: Perspectives and challenges},
  author={Zha, Daochen and Bhat, Zaid Pervaiz and Lai, Kwei-Herng and Yang, Fan and Hu, Xia},
  booktitle={Proceedings of the 2023 SIAM international conference on data mining (SDM)},
  pages={945--948},
  year={2023},
  organization={SIAM}
}

@article{gadre2023datacomp,
  title={Datacomp: In search of the next generation of multimodal datasets},
  author={Gadre, Samir Yitzhak and Ilharco, Gabriel and Fang, Alex and Hayase, Jonathan and Smyrnis, Georgios and Nguyen, Thao and Marten, Ryan and Wortsman, Mitchell and Ghosh, Dhruba and Zhang, Jieyu and others},
  journal={Advances in Neural Information Processing Systems},
  volume={36},
  pages={27092--27112},
  year={2023}
}

@article{zhou2024lima,
  title={Lima: Less is more for alignment},
  author={Zhou, Chunting and Liu, Pengfei and Xu, Puxin and Iyer, Srinivasan and Sun, Jiao and Mao, Yuning and Ma, Xuezhe and Efrat, Avia and Yu, Ping and Yu, Lili and others},
  journal={Advances in Neural Information Processing Systems},
  volume={36},
  pages={55006--55021},
  year={2023}
}

@article{docgenome,
  title={Docgenome: An open large-scale scientific document benchmark for training and testing multi-modal large language models},
  author={Xia, Renqiu and Mao, Song and Yan, Xiangchao and Zhou, Hongbin and Zhang, Bo and Peng, Haoyang and Pi, Jiahao and Fu, Daocheng and Wu, Wenjie and Ye, Hancheng and others},
  journal={arXiv preprint arXiv:2406.11633},
  year={2024}
}

@inproceedings{seung1992qbc,
  title={Query by committee},
  author={Seung, H Sebastian and Opper, Manfred and Sompolinsky, Haim},
  booktitle={Proceedings of the fifth annual workshop on Computational learning theory},
  pages={287--294},
  year={1992}
}

@article{freund1997qbc,
  title={Selective sampling using the query by committee algorithm},
  author={Freund, Yoav and Seung, H Sebastian and Shamir, Eli and Tishby, Naftali},
  journal={Machine learning},
  volume={28},
  number={2},
  pages={133--168},
  year={1997},
  publisher={Springer}
}

@inproceedings{levenshtein1966,
  title={Binary codes capable of correcting deletions, insertions, and reversals},
  author={Levenshtein, Vladimir I and others},
  booktitle={Soviet physics doklady},
  volume={10},
  number={8},
  pages={707--710},
  year={1966},
  organization={Soviet Union}
}

@inproceedings{zhong2020teds,
  title={Image-based table recognition: data, model, and evaluation},
  author={Zhong, Xu and ShafieiBavani, Elaheh and Jimeno Yepes, Antonio},
  booktitle={European conference on computer vision},
  pages={564--580},
  year={2020},
  organization={Springer}
}

@inproceedings{wang2025cdm,
  title={Image over text: Transforming formula recognition evaluation with character detection matching},
  author={Wang, Bin and Wu, Fan and Ouyang, Linke and Gu, Zhuangcheng and Zhang, Rui and Xia, Renqiu and Shi, Botian and Zhang, Bo and He, Conghui},
  booktitle={Proceedings of the Computer Vision and Pattern Recognition Conference},
  pages={19681--19690},
  year={2025}
}

@inproceedings{omnidocbench,
  title={Omnidocbench: Benchmarking diverse pdf document parsing with comprehensive annotations},
  author={Ouyang, Linke and Qu, Yuan and Zhou, Hongbin and Zhu, Jiawei and Zhang, Rui and Lin, Qunshu and Wang, Bin and Zhao, Zhiyuan and Jiang, Man and Zhao, Xiaomeng and others},
  booktitle={Proceedings of the IEEE/CVF Conference on Computer Vision and Pattern Recognition},
  pages={24838--24848},
  year={2025}
}

@article{ocrbench,
  title={Ocrbench: on the hidden mystery of ocr in large multimodal models},
  author={Liu, Yuliang and Li, Zhang and Huang, Mingxin and Yang, Biao and Yu, Wenwen and Li, Chunyuan and Yin, Xu-Cheng and Liu, Cheng-Lin and Jin, Lianwen and Bai, Xiang},
  journal={Science China Information Sciences},
  volume={67},
  number={12},
  pages={220102},
  year={2024},
  publisher={Springer}
}

@inproceedings{cc_ocr,
  title={Cc-ocr: A comprehensive and challenging ocr benchmark for evaluating large multimodal models in literacy},
  author={Yang, Zhibo and Tang, Jun and Li, Zhaohai and Wang, Pengfei and Wan, Jianqiang and Zhong, Humen and Liu, Xuejing and Yang, Mingkun and Wang, Peng and Bai, Shuai and others},
  booktitle={Proceedings of the IEEE/CVF International Conference on Computer Vision},
  pages={21744--21754},
  year={2025}
}

@article{shao2024grpo,
  title={Deepseekmath: Pushing the limits of mathematical reasoning in open language models},
  author={Shao, Zhihong and Wang, Peiyi and Zhu, Qihao and Xu, Runxin and Song, Junxiao and Bi, Xiao and Zhang, Haowei and Zhang, Mingchuan and Li, YK and Wu, Yang and others},
  journal={arXiv preprint arXiv:2402.03300},
  year={2024}
}

@article{yu2025dapo,
  title={Dapo: An open-source llm reinforcement learning system at scale},
  author={Yu, Qiying and Zhang, Zheng and Zhu, Ruofei and Yuan, Yufeng and Zuo, Xiaochen and Yue, Yu and Dai, Weinan and Fan, Tiantian and Liu, Gaohong and Liu, Lingjun and others},
  journal={arXiv preprint arXiv:2503.14476},
  year={2025}
}

@misc{yin2026youtuparsingperceptionstructuringrecognition,
      title={Youtu-Parsing: Perception, Structuring and Recognition via High-Parallelism Decoding}, 
      author={Kun Yin and Yunfei Wu and Bing Liu and Zhongpeng Cai and Xiaotian Li and Huang Chen and Xin Li and Haoyu Cao and Yinsong Liu and Deqiang Jiang and Xing Sun and Yunsheng Wu and Qianyu Li and Antai Guo and Yanzhen Liao and Yanqiu Qu and Haodong Lin and Chengxu He and Shuangyin Liu},
      year={2026},
      eprint={2601.20430},
      archivePrefix={arXiv},
      primaryClass={cs.CV},
      url={https://arxiv.org/abs/2601.20430}, 
}

@misc{chen2025logicsparsingtechnicalreport,
      title={Logics-Parsing Technical Report}, 
      author={Xiangyang Chen and Shuzhao Li and Xiuwen Zhu and Yongfan Chen and Fan Yang and Cheng Fang and Lin Qu and Xiaoxiao Xu and Hu Wei and Minggang Wu},
      year={2025},
      eprint={2509.19760},
      archivePrefix={arXiv},
      primaryClass={cs.CV},
      url={https://arxiv.org/abs/2509.19760}, 
}

@misc{wu2026fireredocrtechnicalreport,
      title={FireRed-OCR Technical Report}, 
      author={Hao Wu and Haoran Lou and Xinyue Li and Zuodong Zhong and Zhaojun Sun and Phellon Chen and Xuanhe Zhou and Kai Zuo and Yibo Chen and Xu Tang and Yao Hu and Boxiang Zhou and Jian Wu and Yongji Wu and Wenxin Yu and Yingmiao Liu and Yuhao Huang and Manjie Xu and Gang Liu and Yidong Ma and Zhichao Sun and Changhao Qiao},
      year={2026},
      eprint={2603.01840},
      archivePrefix={arXiv},
      primaryClass={cs.CV},
      url={https://arxiv.org/abs/2603.01840}, 
}

@article{du2025unirec,
  title={UniRec-0.1B: Unified Text and Formula Recognition with 0.1B Parameters},
  author={Yongkun Du and Zhineng Chen and Yazhen Xie and Weikang Bai and Hao Feng and Wei Shi and Yuchen Su and Can Huang and Yu-Gang Jiang},
  journal={arXiv preprint arXiv:2512.21095},
  year={2025}
}

@inproceedings{dolphin,
  title={Dolphin: Document Image Parsing via Heterogeneous Anchor Prompting},
  author={Feng, Hao and Wei, Shu and Fei, Xiang and Shi, Wei and Han, Yingdong and Liao, Lei and Lu, Jinghui and Wu, Binghong and Liu, Qi and Lin, Chunhui and Tang, Jingqun and Liu, Hao and Huang, Can},
  booktitle={Proceedings of the 65th Annual Meeting of the Association for Computational Linguistics (ACL)},
  year={2025}
}

@misc{zhong2026ocrverseholisticocrendtoend,
      title={OCRVerse: Towards Holistic OCR in End-to-End Vision-Language Models},
      author={Yufeng Zhong and Lei Chen and Xuanle Zhao and Wenkang Han and Liming Zheng and Jing Huang and Deyang Jiang and Yilin Cao and Lin Ma and Zhixiong Zeng},
      year={2026},
      eprint={2601.21639},
      archivePrefix={arXiv},
      primaryClass={cs.CV},
      url={https://arxiv.org/abs/2601.21639}, 
}

@misc{lu2024ovisstructuralembeddingalignment,
      title={Ovis: Structural Embedding Alignment for Multimodal Large Language Model}, 
      author={Shiyin Lu and Yang Li and Qing-Guo Chen and Zhao Xu and Weihua Luo and Kaifu Zhang and Han-Jia Ye},
      year={2024},
      eprint={2405.20797},
      archivePrefix={arXiv},
      primaryClass={cs.CV},
      url={https://arxiv.org/abs/2405.20797}, 
}

@misc{yang2025qwen3technicalreport,
      title={Qwen3 Technical Report},
      author={An Yang and Anfeng Li and Baosong Yang and Beichen Zhang and Binyuan Hui and Bo Zheng and Bowen Yu and Chang Gao and Chengen Huang and Chenxu Lv and Chujie Zheng and Dayiheng Liu and Fan Zhou and Fei Huang and Feng Hu and Hao Ge and Haoran Wei and Huan Lin and Jialong Tang and Jian Yang and Jianhong Tu and Jianwei Zhang and Jianxin Yang and Jiaxi Yang and Jing Zhou and Jingren Zhou and Junyang Lin and Kai Dang and Keqin Bao and Kexin Yang and Le Yu and Lianghao Deng and Mei Li and Mingfeng Xue and Mingze Li and Pei Zhang and Peng Wang and Qin Zhu and Rui Men and Ruize Gao and Shixuan Liu and Shuang Luo and Tianhao Li and Tianyi Tang and Wenbiao Yin and Xingzhang Ren and Xinyu Wang and Xinyu Zhang and Xuancheng Ren and Yang Fan and Yang Su and Yichang Zhang and Yinger Zhang and Yu Wan and Yuqiong Liu and Zekun Wang and Zeyu Cui and Zhenru Zhang and Zhipeng Zhou and Zihan Qiu},
      year={2025},
      eprint={2505.09388},
      archivePrefix={arXiv},
      primaryClass={cs.CL},
      url={https://arxiv.org/abs/2505.09388}, 
}

@misc{wang2025internvl35advancingopensourcemultimodal,
      title={InternVL3.5: Advancing Open-Source Multimodal Models in Versatility, Reasoning, and Efficiency}, 
      author={Weiyun Wang and Zhangwei Gao and Lixin Gu and Hengjun Pu and Long Cui and Xingguang Wei and Zhaoyang Liu and Linglin Jing and Shenglong Ye and Jie Shao and Zhaokai Wang and Zhe Chen and Hongjie Zhang and Ganlin Yang and Haomin Wang and Qi Wei and Jinhui Yin and Wenhao Li and Erfei Cui and Guanzhou Chen and Zichen Ding and Changyao Tian and Zhenyu Wu and Jingjing Xie and Zehao Li and Bowen Yang and Yuchen Duan and Xuehui Wang and Zhi Hou and Haoran Hao and Tianyi Zhang and Songze Li and Xiangyu Zhao and Haodong Duan and Nianchen Deng and Bin Fu and Yinan He and Yi Wang and Conghui He and Botian Shi and Junjun He and Yingtong Xiong and Han Lv and Lijun Wu and Wenqi Shao and Kaipeng Zhang and Huipeng Deng and Biqing Qi and Jiaye Ge and Qipeng Guo and Wenwei Zhang and Songyang Zhang and Maosong Cao and Junyao Lin and Kexian Tang and Jianfei Gao and Haian Huang and Yuzhe Gu and Chengqi Lyu and Huanze Tang and Rui Wang and Haijun Lv and Wanli Ouyang and Limin Wang and Min Dou and Xizhou Zhu and Tong Lu and Dahua Lin and Jifeng Dai and Weijie Su and Bowen Zhou and Kai Chen and Yu Qiao and Wenhai Wang and Gen Luo},
      year={2025},
      eprint={2508.18265},
      archivePrefix={arXiv},
      primaryClass={cs.CV},
      url={https://arxiv.org/abs/2508.18265}, 
}

@misc{qwen35blog,
    title = {Qwen3.5: Accelerating Productivity with Native Multimodal Agents},
    url = {https://qwen.ai/blog?id=qwen3.5},
    author = {Qwen Team},
    month = {February},
    year = {2026}
}

@article{lu2025ovis25technicalreport,
  title={Ovis2.5 Technical Report}, 
  author={Shiyin Lu and Yang Li and Yu Xia and Yuwei Hu and Shanshan Zhao and Yanqing Ma and Zhichao Wei and Yinglun Li and Lunhao Duan and Jianshan Zhao and Yuxuan Han and Haijun Li and Wanying Chen and Junke Tang and Chengkun Hou and Zhixing Du and Tianli Zhou and Wenjie Zhang and Huping Ding and Jiahe Li and Wen Li and Gui Hu and Yiliang Gu and Siran Yang and Jiamang Wang and Hailong Sun and Yibo Wang and Hui Sun and Jinlong Huang and Yuping He and Shengze Shi and Weihong Zhang and Guodong Zheng and Junpeng Jiang and Sensen Gao and Yi-Feng Wu and Sijia Chen and Yuhui Chen and Qing-Guo Chen and Zhao Xu and Weihua Luo and Kaifu Zhang},
  year={2025},
  journal={arXiv:2508.11737}
}

@article{wang2024unimernet,
  title={Unimernet: A universal network for real-world mathematical expression recognition},
  author={Wang, Bin and Gu, Zhuangcheng and Liang, Guang and Xu, Chao and Zhang, Bo and Shi, Botian and He, Conghui},
  journal={arXiv preprint arXiv:2404.15254},
  year={2024}
}

@article{zhao2024doclayout,
  title={Doclayout-yolo: Enhancing document layout analysis through diverse synthetic data and global-to-local adaptive perception},
  author={Zhao, Zhiyuan and Kang, Hengrui and Wang, Bin and He, Conghui},
  journal={arXiv preprint arXiv:2410.12628},
  year={2024}
}

\clearpage

\beginappendix

\section{Prompt Design and Task Examples}
\label{app:prompt_design}

This section provides the prompt formats, output specifications, and representative examples for each task supported by \minerupro. All tasks share a unified prompt interface: a single \texttt{<image>} token followed by a plain-text task suffix, requiring no few-shot examples or structured metadata.

\paragraph{Task Prompt Convention.}
The five task suffixes and their output formats are summarized below:

\begin{itemize}
    \item \textbf{Layout Detection} (\S\ref{app:layout_detection}) — localizes content regions and outputs bounding boxes with category labels and rotation flags.
    \item \textbf{Text Recognition} (\S\ref{app:text_recognition}) — transcribes cropped text regions into plain text.
    \item \textbf{Formula Recognition} (\S\ref{app:formula_recognition}) — converts cropped formula regions into \LaTeX{} markup.
    \item \textbf{Table Recognition} (\S\ref{app:table_recognition}) — serializes cropped tables into an OTSL-based token sequence with cell content, subsequently converted to HTML.
    \item \textbf{Image Analysis} (\S\ref{app:image_aware}) — classifies image regions and extracts captions and embedded content.
\end{itemize}

\begin{figure*}[!htbp]
\centering
\includegraphics[width=0.92\textwidth]{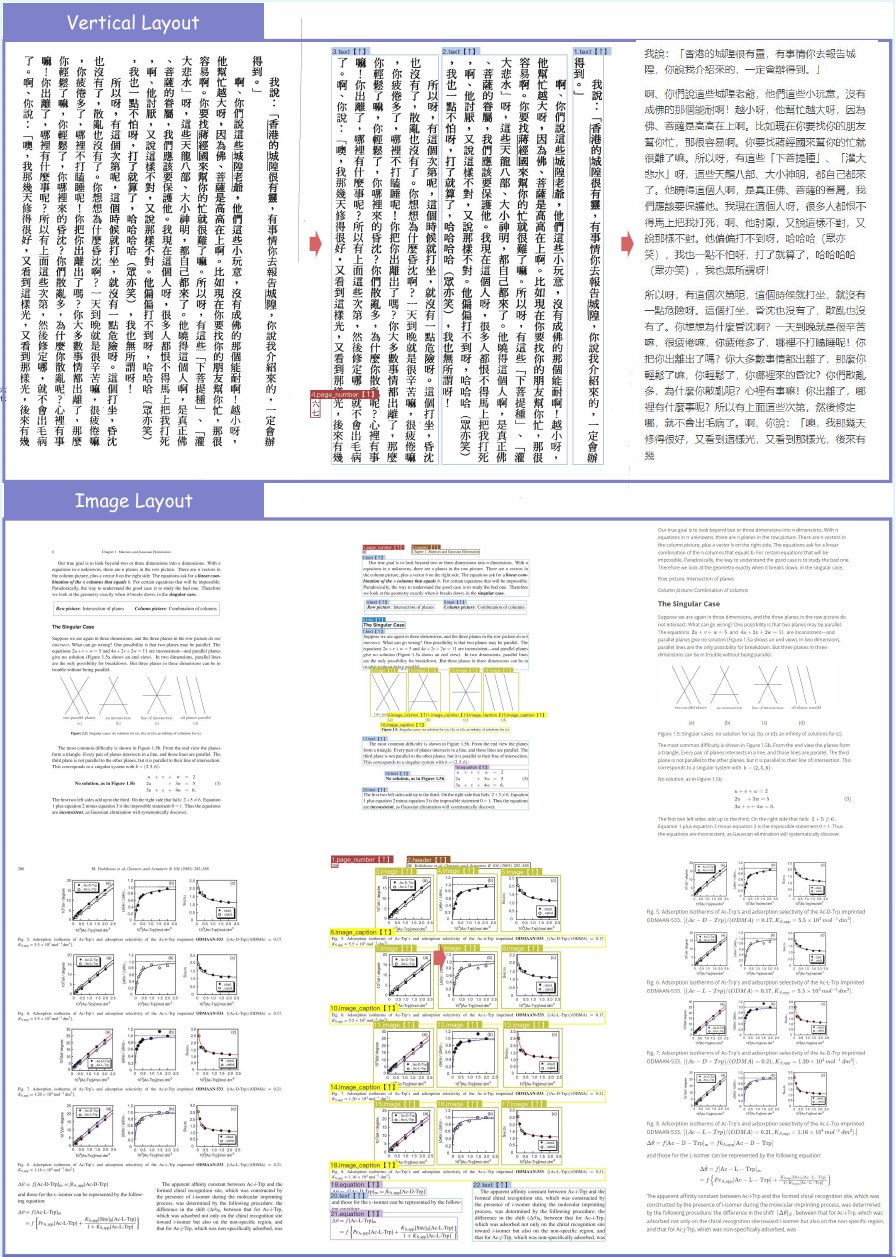}
\caption{Layout Detection examples. The model localizes content regions with bounding boxes, category labels, and rotation flags on diverse document pages.}
\label{fig:layout_examples}
\end{figure*}

\subsection{Layout Detection}
\label{app:layout_detection}

Layout Detection serves as the entry point of the document parsing pipeline, responsible for localizing all content regions on a page and assigning each a semantic category. The model takes a downsampled page image and produces a sequence of structured region descriptors.

\paragraph{Prompt.}
\begin{verbatim}
<image>\nLayout Detection:
\end{verbatim}

\paragraph{Output Format.}
The output is a newline-delimited sequence of region descriptors, where each region follows the format:

\begin{verbatim}
<|box_start|>x1 y1 x2 y2<|box_end|><|ref_start|>category<|ref_end|><|rotate_dir|>
\end{verbatim}

\noindent Here, \texttt{x1 y1 x2 y2} are the normalized bounding box coordinates (scaled to a $[0, 999]$ grid), \texttt{category} is the semantic label of the region (e.g., \texttt{title}, \texttt{text}, \texttt{header}, \texttt{footer}, \texttt{table}, \texttt{figure}, \texttt{formula}), and \texttt{<|rotate\_dir|>} indicates the text orientation (\texttt{<|rotate\_up|>} for standard upright text, with other directions for rotated content). Regions are emitted in natural reading order (top-to-bottom, left-to-right for left-to-right scripts).

\paragraph{Example.}
Given a document title page, the model outputs:
\begin{verbatim}
<|box_start|>705 112 899 146<|box_end|><|ref_start|>header<|ref_end|><|rotate_up|>
<|box_start|>030 343 132 397<|box_end|><|ref_start|>title<|ref_end|><|rotate_up|>
<|box_start|>212 330 491 382<|box_end|><|ref_start|>title<|ref_end|><|rotate_up|>
<|box_start|>214 389 767 441<|box_end|><|ref_start|>title<|ref_end|><|rotate_up|>
<|box_start|>219 494 359 523<|box_end|><|ref_start|>text<|ref_end|><|rotate_up|>
<|box_start|>654 940 907 975<|box_end|><|ref_start|>footer<|ref_end|><|rotate_up|>
\end{verbatim}

\noindent This output identifies six regions—one header, three title blocks, one text block, and one footer—each with precise spatial coordinates and upright orientation. Additional examples are provided in \Cref{fig:layout_examples}.


\subsection{Text Recognition}
\label{app:text_recognition}

Text Recognition transcribes cropped text regions into plain text. Each region is an original-resolution crop produced by Stage~1 Layout Detection.

\paragraph{Prompt.}
\begin{verbatim}
<image>\nText Recognition:
\end{verbatim}

\begin{figure*}[!htbp]
\centering
\includegraphics[width=0.90\textwidth]{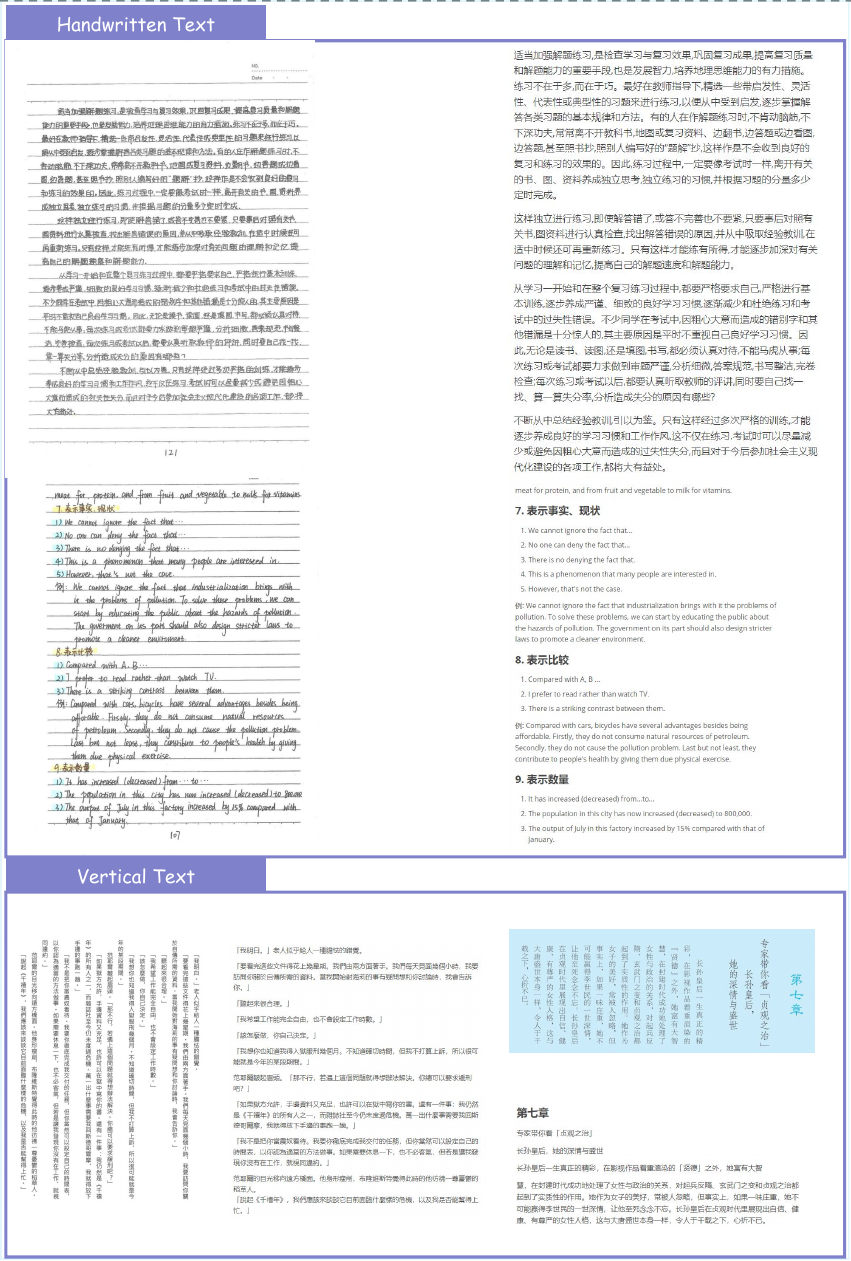}
\caption{Text Recognition examples across Chinese, English, and mixed-language text regions.}
\label{fig:text_rec_examples}
\end{figure*}

\paragraph{Output Format.}
The output is a plain-text string corresponding to the content of the cropped text region. No special tokens or markup are used—the model generates raw text as-is, including whitespace, punctuation, and any inline symbols present in the source image. Additional examples are provided in \Cref{fig:text_rec_examples}.

\begin{figure*}[!htbp]
\centering
\includegraphics[width=0.90\textwidth]{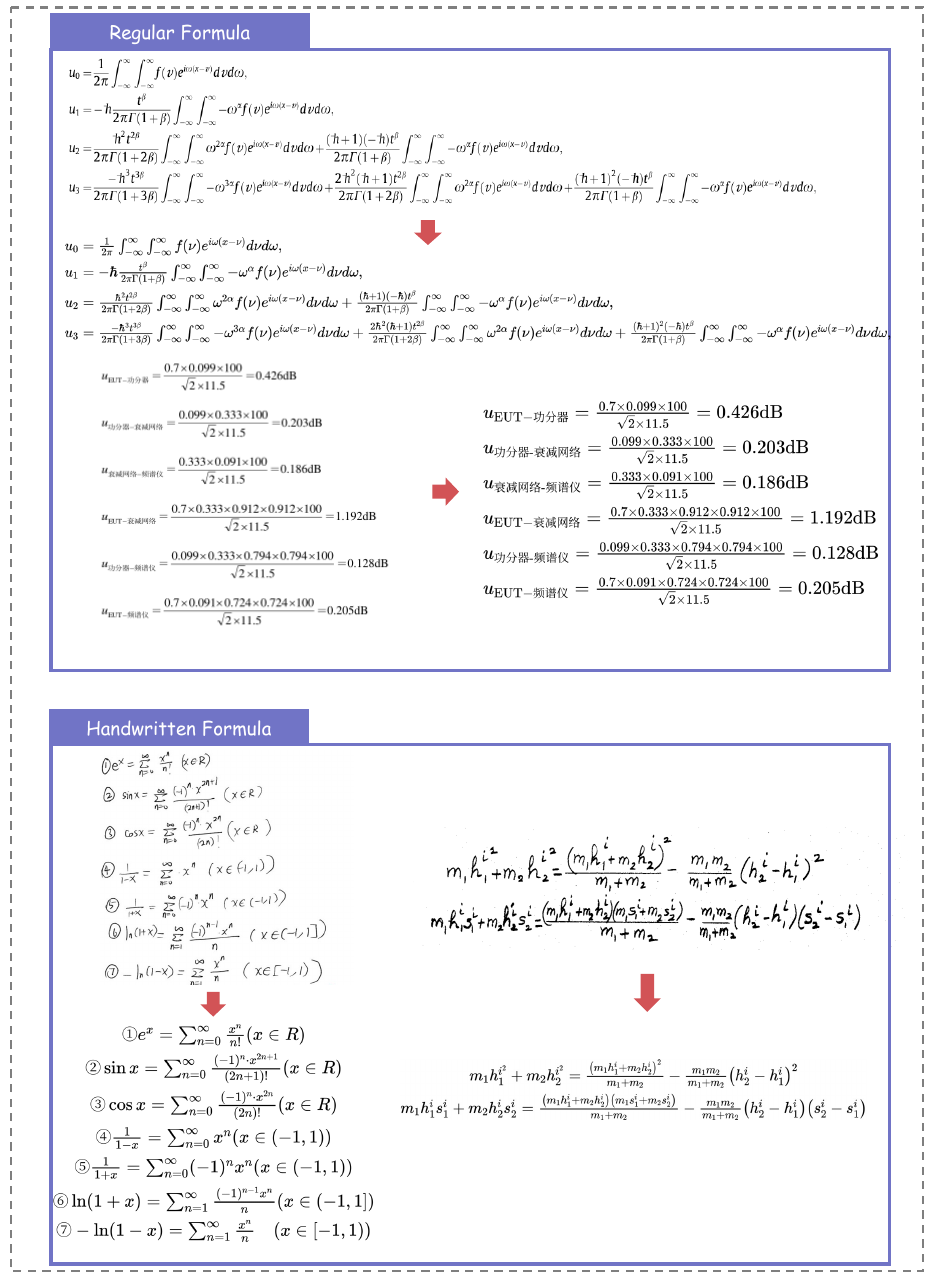}
\caption{Formula Recognition examples including single-line display formulas and complex multi-line equations.}
\label{fig:formula_rec_examples}
\end{figure*}


\subsection{Formula Recognition}
\label{app:formula_recognition}

Formula Recognition converts cropped formula regions into \LaTeX{} markup. The model supports both inline and display-style formulas, as well as multi-line equation environments.

\paragraph{Prompt.}
\begin{verbatim}
<image>\nFormula Recognition:
\end{verbatim}

\paragraph{Output Format.}
The output is a \LaTeX{} math string. Display-style (block) formulas are wrapped in \verb|\[…\]| delimiters. Equation numbers, when present in the source image, are preserved via \verb|\tag{…}|. The model generates standard \LaTeX{} math commands and environments (e.g., \verb|\frac|, \verb|\mathrm|, \verb|\quad|), ensuring the output is directly compilable.

\paragraph{Example: Single Display Formula.}
Given a cropped equation region, the model outputs:
\begin{verbatim}
\[L = 0.004 \ln (2D/d) \quad (\mu \mathrm{H} / \mathrm{cm}) \tag{4-2-10}\]
\end{verbatim}

\noindent The equation number \texttt{(4-2-10)} from the original document is captured via \verb|\tag{}|.

\paragraph{Multi-line Formulas.}
Multi-line equations are handled through the collaboration of Layout Detection and Formula Recognition. Layout Detection first identifies an \texttt{equation\_block} region encompassing the entire multi-line group, within which individual single-line formulas are separately localized. Each line is then independently cropped and recognized by Formula Recognition. The final multi-line output is produced by concatenating the individual \LaTeX{} results in reading order, faithfully reproducing the original equation group without requiring the model to generate multi-line environments in a single pass.

Additional examples are provided in \Cref{fig:formula_rec_examples}.


\subsection{Table Recognition}
\label{app:table_recognition}

Table Recognition converts cropped table regions into a structured token sequence based on OTSL (Optimized Table Structure Language). Cell content is transcribed as plain text, with inline formulas in \LaTeX{} when present. The OTSL output is subsequently converted to HTML for downstream consumption.

\paragraph{Prompt.}
\begin{verbatim}
<image>\nTable Recognition:
\end{verbatim}

\paragraph{Output Format.}
The output is a flat token sequence representing the table structure row by row. Each cell is delimited by \texttt{<fcel>}, and rows are separated by \texttt{<nl>}. Cell content may contain plain text, \LaTeX{} inline math (\verb|\(…\)|), or a mixture of both. The OTSL representation is compact and unambiguous, supporting regular grids as well as cells with complex content. After generation, the OTSL sequence is programmatically converted to HTML for rendering and downstream integration.

\paragraph{Example.}
Given a cropped table with two rows (a header row of time values and a data row of concentration values), the model outputs:

\begin{verbatim}
<fcel>\( \frac{t}{\min} \)<fcel>3<fcel>5<fcel>7<fcel>10<fcel>15<fcel>21<fcel>25<nl>
<fcel>\( \frac{1/c}{{\mathrm{\;{mol}}}^{-1}\cdot {\mathrm{{dm}}}^{3}} \)<fcel>135.1
<fcel>157.7<fcel>181.8<fcel>215.5<fcel>275.5<fcel>347.2<fcel>393.2<nl>
\end{verbatim}

\noindent Each \texttt{<fcel>} token introduces a cell, and \texttt{<nl>} marks the end of a row. The first column contains \LaTeX{}-formatted headers with units, while the remaining columns hold numeric values.

\begin{figure*}[!htbp]
\centering
\includegraphics[width=0.90\textwidth]{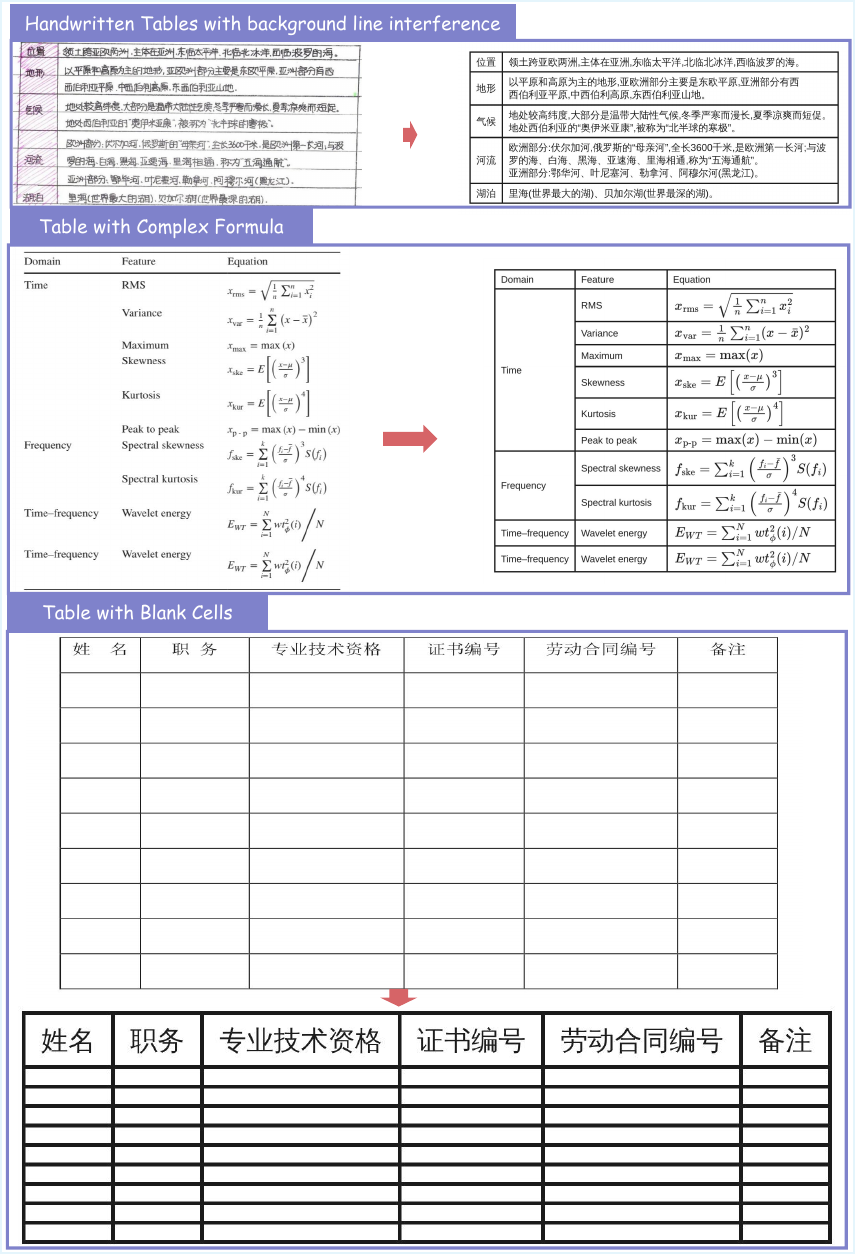}
\caption{Table Recognition examples showing OTSL token output and the corresponding rendered HTML for tables with varying complexity.}
\label{fig:table_rec_examples}
\end{figure*}

Additional examples are provided in \Cref{fig:table_rec_examples}.


\subsection{Image-Aware Parsing}
\label{app:image_aware}

\begin{figure*}[!htbp]
\centering
\includegraphics[width=0.90\textwidth]{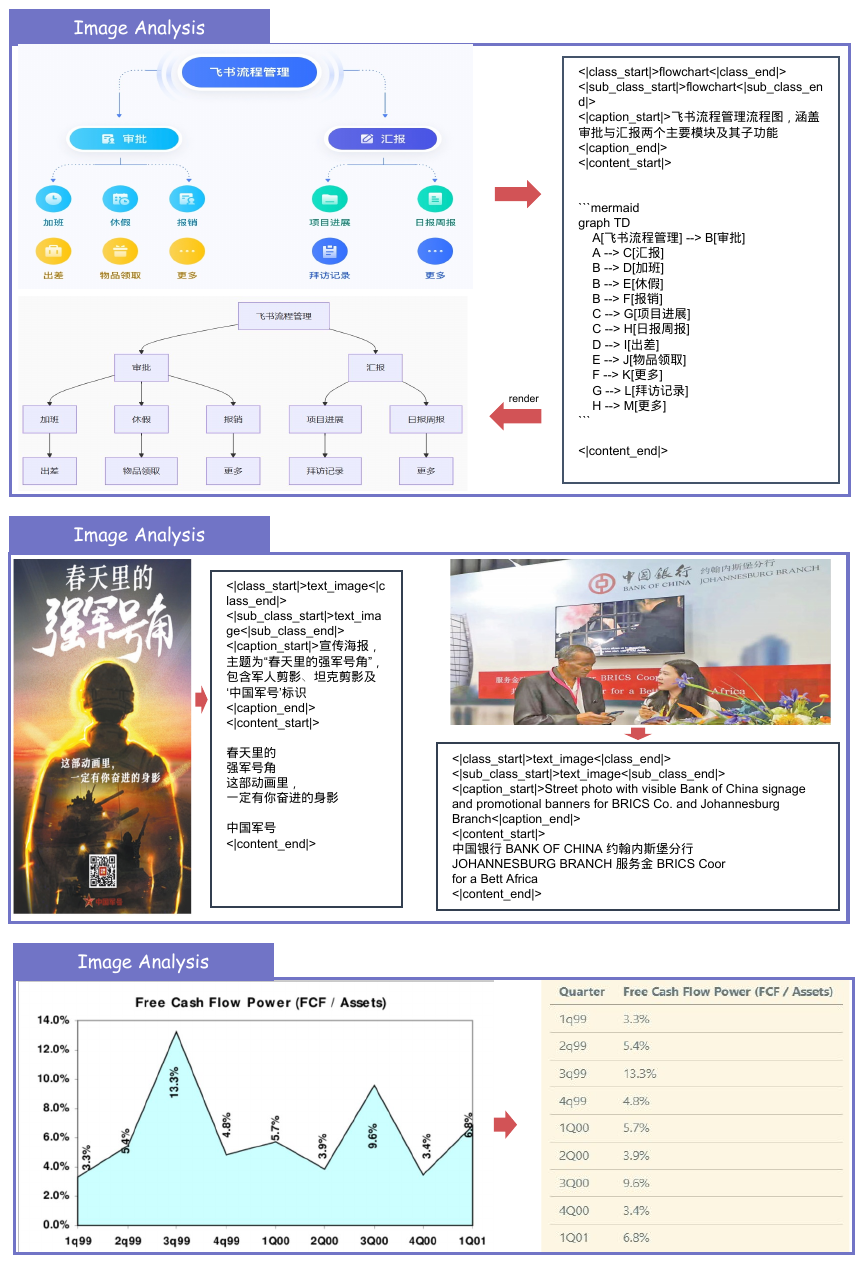}
\caption{Image-aware parsing examples. The model classifies image regions into fine-grained subtypes and extracts structured content accordingly.}
\label{fig:image_parsing}
\end{figure*}

Image Analysis classifies cropped image regions and extracts their embedded content. Unlike other recognition tasks that target a single modality, Image Analysis first determines the semantic type of the image and then extracts structured content accordingly—text, formulas, tables, or a combination thereof.

\paragraph{Prompt.}
\begin{verbatim}
<image>\nImage Analysis:
\end{verbatim}

\paragraph{Output Format.}
The output consists of four structured fields delimited by special tokens:

\begin{verbatim}
<|class_start|>class<|class_end|>
<|sub_class_start|>sub_class<|sub_class_end|>
<|caption_start|>caption<|caption_end|>
<|content_start|>content<|content_end|>
\end{verbatim}

\noindent Here, \texttt{class} is the primary image category (e.g., \texttt{pure\_formula}, \texttt{natural\_image}, \texttt{chart}), \texttt{sub\_class} provides a finer-grained label, \texttt{caption} captures any associated caption text (left empty if absent), and \texttt{content} contains the extracted textual or structured content from within the image.

\paragraph{Example.}
Given a cropped figure region containing a standalone formula, the model outputs:
\begin{verbatim}
<|class_start|>pure_formula<|class_end|>
<|sub_class_start|>pure_formula<|sub_class_end|>
<|caption_start|><|caption_end|>
<|content_start|>p + q = 1<|content_end|>
\end{verbatim}

\noindent The image is classified as \texttt{pure\_formula} with no caption, and the formula content is directly extracted.
Additional examples are provided in \Cref{fig:image_parsing}.

\section{Extended Parsing Capabilities}
\label{app:system_enhancements}

Beyond improvements in recognition accuracy, \minerupro extends the parsing capabilities of \mineru in several practical dimensions. These features target real-world deployment scenarios where documents are multi-page, richly illustrated, and structurally complex. While they do not affect OmniDocBench scores (which focus on single-page content recognition), they substantially improve end-to-end parsing completeness and usability.

\paragraph{Image-aware parsing.}
\mineru crops all image regions without further processing, discarding potentially valuable information such as chart data, embedded text, and diagram content. \minerupro introduces image-aware parsing (\S\ref{app:image_aware}) that first classifies each image region into fine-grained subtypes (chart, text image, table-like image, general image) and then applies differentiated extraction strategies: charts are parsed into structured tables, text images undergo OCR, and table-like images are recognized as tables. This framework is readily extensible to additional image types; however, we have not yet applied Data Engine optimization to image analysis data in this release, leaving significant room for future improvement.

\paragraph{Truncated paragraph merging.}

\begin{figure*}[!htbp]
\centering
\includegraphics[width=0.90\textwidth]{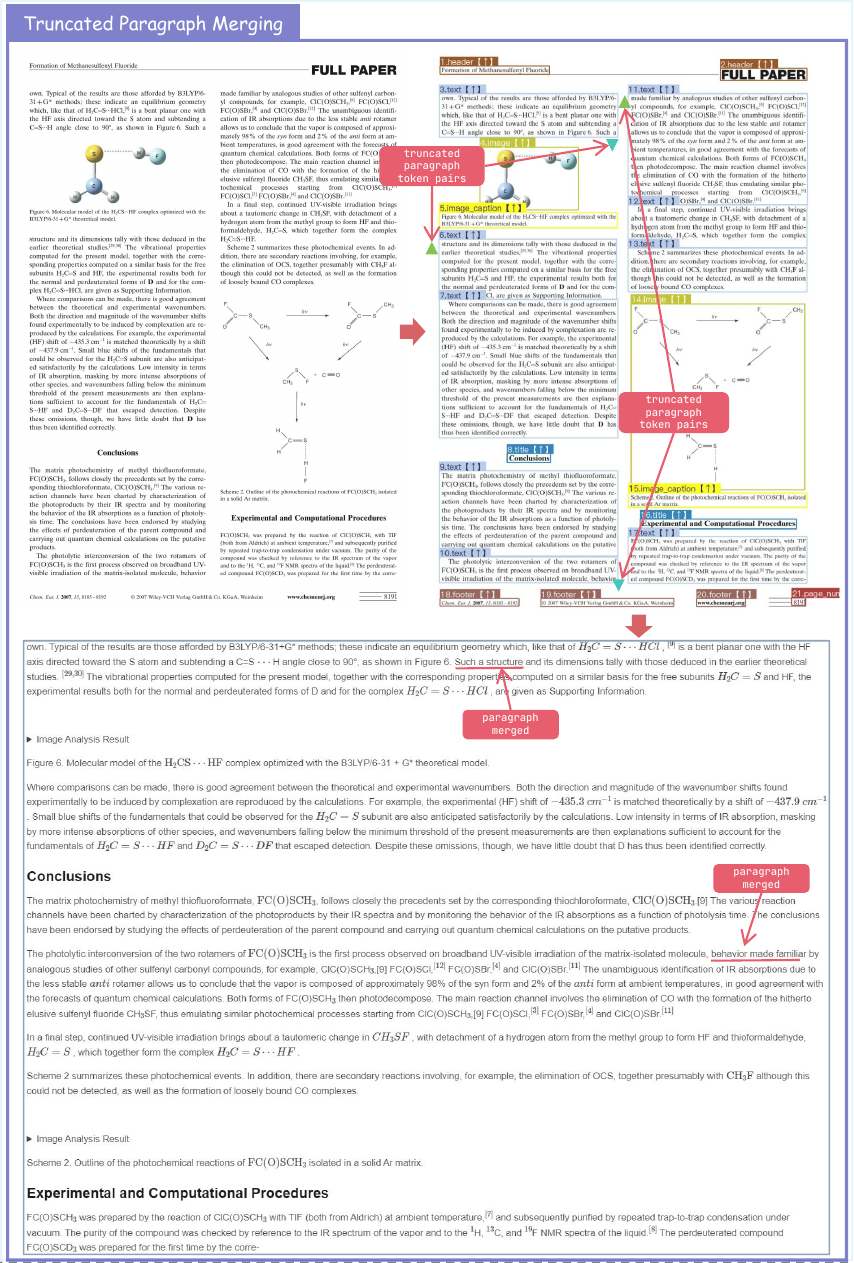}
\caption{Illustration of truncated paragraph merging. In multi-column and complex layouts, Layout Detection splits continuous paragraphs into separate bounding boxes. The merge label predicted by the model reassembles them into coherent paragraphs during Markdown rendering.}
\label{fig:truncated_paragraph_merge}
\end{figure*}

Layout Detection tends to segment each spatially distinct text block as an independent region, which can split semantically continuous paragraphs into multiple fragments. Common causes include column boundaries in multi-column layouts, figures or tables interrupting a paragraph, and unusually wide line spacing. To address this, \minerupro performs \emph{truncated paragraph merging} as part of the Layout Detection task. Since Layout Detection already establishes reading order, and truncation necessarily occurs between consecutive regions in that order, the problem reduces to a binary classification at each adjacent-region boundary: \emph{merge} or \emph{no merge}. This binary label is integrated directly into the layout output sequence, allowing truncated paragraphs to be reassembled during final Markdown rendering without affecting downstream recognition tasks. The merging process is illustrated in \Cref{fig:truncated_paragraph_merge}.

To construct training data for this capability, we annotate merge decisions on top of existing layout ground truth. For each pair of adjacent \texttt{text} or \texttt{list\_item} regions, we first apply rule-based filtering using sentence length, leading numbering patterns, and terminal punctuation to eliminate obvious non-merge cases. For the remaining candidates, we highlight the two regions in red and green on the page image and query Gemini~3~Flash with both the annotated image and the text content of each region, asking it to judge whether merging is appropriate based on layout context and textual coherence. To reduce API cost, only the first and last sentences are provided for long paragraphs.

\paragraph{Cross-page table merging.}

\begin{figure*}[!htbp]
\centering
\includegraphics[width=0.92\textwidth]{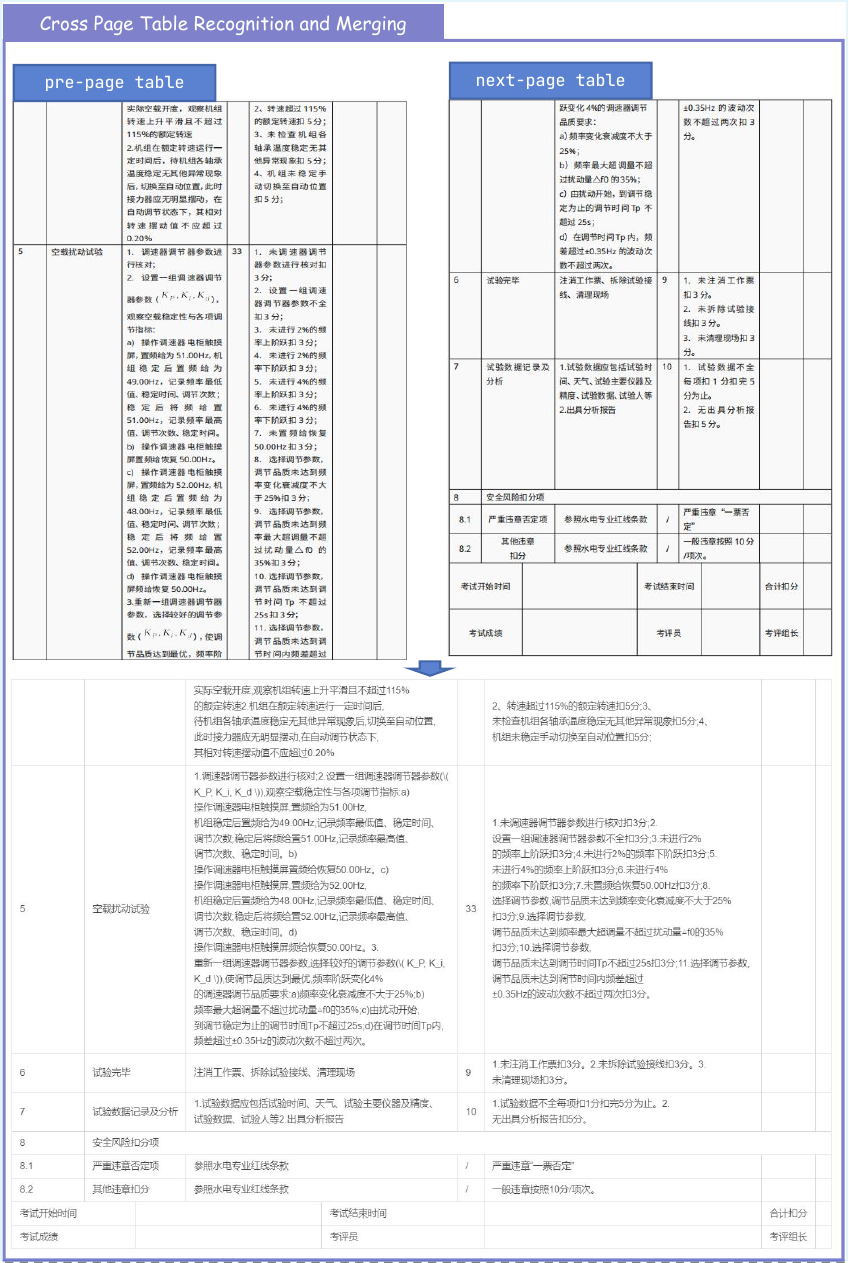}
\caption{Cross-page table merging example. The model performs semantic understanding at the junction of two table fragments identified by rule-based heuristics, producing per-column merge decisions to reconstruct the complete table.}
\label{fig:cross_page_table}
\end{figure*}


When a table is split across a page break, \minerupro automatically detects and merges the fragments. The system first applies rule-based heuristics to identify candidate pairs: if the last table on a page and the first table on the next page share compatible column counts and structural patterns, they are flagged for merging. For flagged pairs, the model receives the last row(s) of the upper table and the first data row(s) of the lower table as a structured text prompt:

\begin{verbatim}
Please merge the next two tables.

## Table 1 (Previous Page - Last Table)
**Last Row(s) Data:**
[[{content of table 1}]]

## Table 2 (Current Page - First Table)
**First Data Row(s):**
[[{content of table 2}]]
\end{verbatim}

\noindent The model outputs a per-column binary decision list indicating whether each column should be \emph{directly concatenated} (0) or \emph{semantically merged} (1). Direct concatenation applies when cells are cleanly split at the page boundary (\eg a single cell's content is broken across two rows), while semantic merging preserves both rows as distinct data. A typical semantic merging process is shown in \Cref{fig:cross_page_table}. This fine-grained, column-level strategy handles the common case where some columns require concatenation and others do not within the same table split.

\paragraph{In-table image detection.}

Tables in real-world documents frequently contain embedded images (e.g., product photos, diagrams, icons). \minerupro detects these through a three-step process:

\begin{enumerate}
    \item \textbf{Detection.} Layout Detection identifies image regions that fall spatially within a table bounding box. Each detected in-table image is replaced with a special placeholder token in the table crop, effectively masking the image region.
    \item \textbf{Recognition.} The masked table image is fed to Table Recognition, which generates the OTSL sequence with placeholder tokens marking the positions of masked images.
    \item \textbf{Restoration.} In the final output, placeholder tokens are resolved back to references to the original image regions, producing HTML table cells that contain \texttt{<img>} tags with unique identifiers linking to the extracted image content blocks.
\end{enumerate}

\begin{figure*}[!htbp]
\centering
\includegraphics[width=0.90\textwidth]{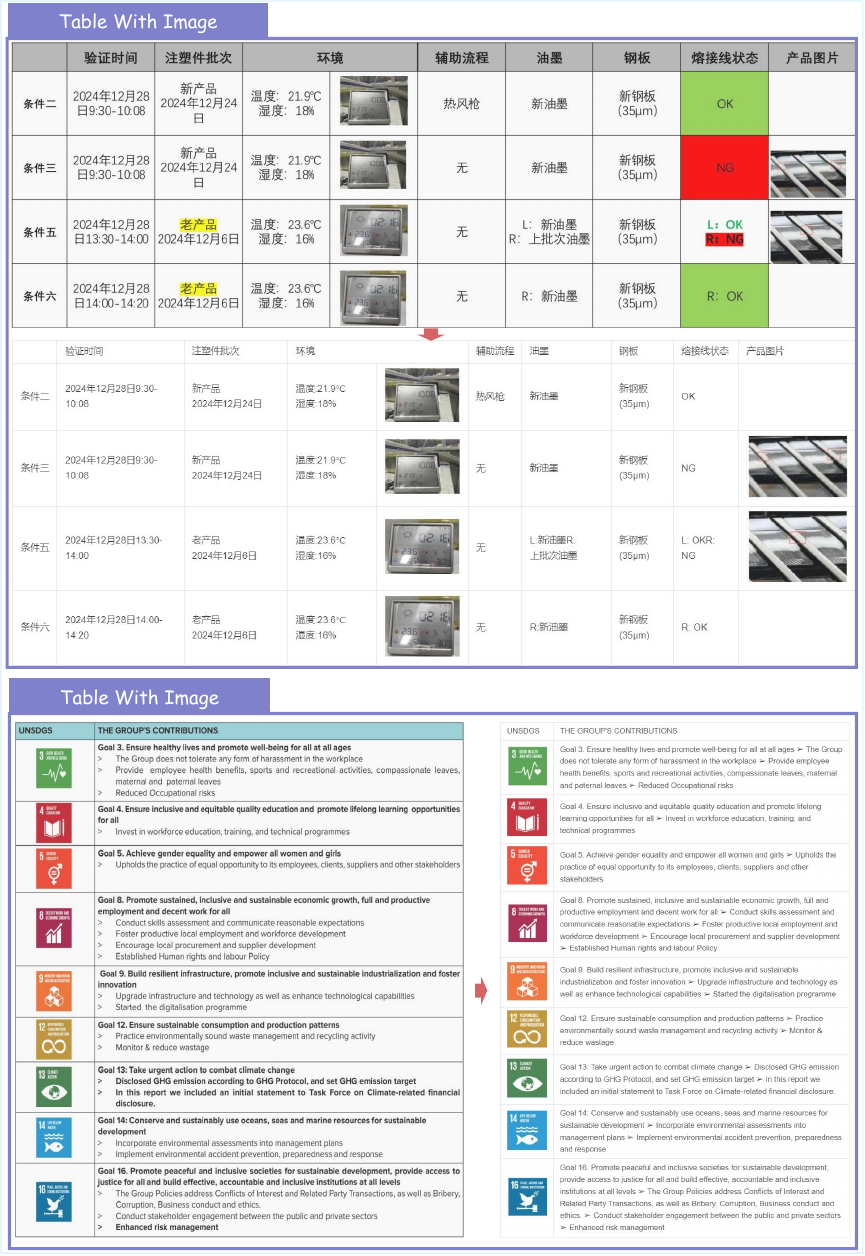}
\caption{In-table image detection and recognition. Embedded images within table cells are masked with placeholders during Table Recognition and restored as \texttt{<img>} references in the final HTML output.}
\label{fig:table_with_image}
\end{figure*}

\noindent This approach allows the table structure and textual content to be recognized without interference from embedded images, while preserving the spatial correspondence between images and their containing cells in the final output. Representative examples are shown in \Cref{fig:table_with_image}.

\section{OmniDocBench v1.6 Detailed Results}
\label{app:omnidocbench_details}

This section reports per-model detailed results on the OmniDocBench~v1.6 Base and Hard subsets. The evaluation protocol and matching corrections are described in \Cref{sec:omnidocbench}.

\subsection{Detailed Results on Base Subset}
\label{app:base_results}

\Cref{tab:base_detailed} reports detailed evaluation results for all models on the OmniDocBench~v1.6 Base subset.

\begin{table*}[t]
\centering
\caption{Detailed results on OmniDocBench~v1.6 Base subset.}
\label{tab:base_detailed}
\small
\setlength{\tabcolsep}{4pt}
\begin{tabular}{lcccccc}
\toprule
\textbf{Model} & \textbf{Overall$\uparrow$} & \textbf{Text Edit$\downarrow$} & \textbf{Formula CDM$\uparrow$} & \textbf{Table TEDS$\uparrow$} & \textbf{TEDS-S$\uparrow$} & \textbf{Read Order$\downarrow$} \\
\midrule
\textbf{GLM-OCR} & \textbf{96.19} & 0.039 & \textbf{98.10} & 94.36 & 96.59 & 0.122 \\
\minerupro & 96.12 & \textbf{0.033} & 97.16 & \textbf{94.49} & \textbf{96.63} & \textbf{0.109} \\
PaddleOCR-VL-1.5 & 95.72 & 0.032 & 97.18 & 93.17 & 95.31 & 0.118 \\
Youtu-Parsing & 94.87 & 0.038 & 94.32 & 94.07 & 96.56 & 0.101 \\
Ovis2.6-30B-A3B & 94.56 & 0.031 & 95.11 & 91.64 & 94.33 & 0.125 \\
PaddleOCR-VL & 94.49 & 0.035 & 95.43 & 91.51 & 94.39 & 0.123 \\
Logics-Parsing-v2 & 94.16 & 0.036 & 95.06 & 91.04 & 93.92 & 0.127 \\
FireRed-OCR & 94.14 & 0.030 & 95.63 & 89.76 & 92.41 & 0.120 \\
\mineru & 93.23 & 0.042 & 94.81 & 89.06 & 92.24 & 0.120 \\
OpenDoc-0.1B & 93.04 & 0.039 & 94.05 & 88.93 & 91.99 & 0.126 \\
Gemini 3 Pro & 92.96 & 0.060 & 95.11 & 89.80 & 93.36 & 0.157 \\
Gemini 3 Flash & 92.58 & 0.062 & 94.19 & 89.74 & 93.82 & 0.163 \\
HunyuanOCR & 92.45 & 0.082 & 91.09 & 94.44 & 95.76 & 0.156 \\
DeepSeek-OCR-2.0 & 91.50 & 0.046 & 93.02 & 86.13 & 89.75 & 0.134 \\
dots.ocr & 90.91 & 0.041 & 88.85 & 88.02 & 90.99 & 0.126 \\
Dolphin-2.0 & 90.42 & 0.064 & 89.98 & 87.67 & 90.31 & 0.137 \\
Qwen3-VL-235B & 90.08 & 0.062 & 91.86 & 84.61 & 87.89 & 0.157 \\
OCRverse & 89.36 & 0.054 & 88.77 & 84.67 & 88.19 & 0.152 \\
MonkeyOCR-pro-3B & 89.15 & 0.067 & 86.87 & 87.22 & 90.49 & 0.181 \\
Dolphin-1.5 & 87.24 & 0.091 & 86.32 & 84.47 & 87.55 & 0.157 \\
GPT-5.2 & 86.83 & 0.120 & 88.62 & 83.82 & 88.31 & 0.188 \\
Mistral-OCR & 86.36 & 0.095 & 89.28 & 79.34 & 83.32 & 0.161 \\
POINTS-Reader & 86.20 & 0.095 & 89.69 & 78.37 & 81.43 & 0.184 \\
Nanonets-OCR-s & 86.10 & 0.099 & 86.43 & 81.75 & 85.41 & 0.192 \\
olmOCR & 85.89 & 0.135 & 87.00 & 84.17 & 87.64 & 0.205 \\
InternVL 3.5 & 83.76 & 0.137 & 89.39 & 75.58 & 80.53 & 0.214 \\
\bottomrule
\end{tabular}
\end{table*}

On the Base subset, top model scores are tightly clustered: the top 6 Overall scores fall within 94.49--96.19, a range of only 1.70 points. \minerupro ranks second at 96.12, only 0.07 points behind GLM-OCR, while leading in Text Edit Distance (0.033 \vs 0.039), Table TEDS (94.49 \vs 94.36), and Reading Order (0.109 \vs 0.122).

\subsection{Detailed Results on Hard Subset}
\label{app:hard_results}

\Cref{tab:hard_detailed} reports detailed evaluation results for all models on the OmniDocBench~v1.6 Hard subset.

\begin{table*}[t]
\centering
\caption{Detailed results on OmniDocBench~v1.6 Hard subset.}
\label{tab:hard_detailed}
\small
\setlength{\tabcolsep}{4pt}
\begin{tabular}{lcccccc}
\toprule
\textbf{Model} & \textbf{Overall$\uparrow$} & \textbf{Text Edit$\downarrow$} & \textbf{Formula CDM$\uparrow$} & \textbf{Table TEDS$\uparrow$} & \textbf{TEDS-S$\uparrow$} & \textbf{Read Order$\downarrow$} \\
\midrule
\textbf{\minerupro} & \textbf{94.08} & \textbf{0.052} & \textbf{97.54} & \textbf{89.91} & \textbf{93.61} & \textbf{0.170} \\
PaddleOCR-VL & 92.48 & 0.066 & 96.24 & 87.84 & 91.60 & 0.189 \\
GLM-OCR & 92.01 & 0.066 & 94.81 & 87.81 & 91.44 & 0.186 \\
PaddleOCR-VL-1.5 & 92.01 & 0.065 & 95.74 & 86.75 & 91.30 & 0.181 \\
Gemini 3 Flash & 91.99 & 0.085 & 96.68 & 87.83 & 92.51 & 0.214 \\
Gemini 3 Pro & 91.99 & 0.083 & 97.23 & 87.03 & 91.68 & 0.198 \\
\mineru & 91.65 & 0.062 & 97.12 & 84.00 & 88.97 & 0.178 \\
Ovis2.6-30B-A3B & 90.39 & 0.056 & 94.56 & 82.21 & 86.07 & 0.184 \\
Logics-Parsing-v2 & 89.95 & 0.062 & 96.28 & 79.81 & 85.61 & 0.184 \\
FireRed-OCR & 89.89 & 0.073 & 94.57 & 82.40 & 86.64 & 0.183 \\
Youtu-Parsing & 89.81 & 0.076 & 91.75 & 85.30 & 89.90 & 0.185 \\
dots.ocr & 88.67 & 0.081 & 89.65 & 84.42 & 89.25 & 0.196 \\
Qwen3-VL-235B & 88.45 & 0.065 & 93.85 & 78.01 & 83.00 & 0.210 \\
DeepSeek-OCR-2.0 & 86.23 & 0.067 & 88.81 & 76.53 & 81.19 & 0.191 \\
GPT-5.2 & 86.07 & 0.087 & 86.79 & 80.10 & 86.70 & 0.213 \\
Dolphin-2.0 & 85.29 & 0.094 & 91.61 & 73.68 & 78.04 & 0.210 \\
MonkeyOCR-pro-3B & 85.07 & 0.109 & 91.18 & 74.92 & 82.48 & 0.228 \\
OCRverse & 84.79 & 0.106 & 89.86 & 75.11 & 79.99 & 0.215 \\
olmOCR & 84.34 & 0.157 & 89.58 & 79.17 & 85.65 & 0.269 \\
InternVL 3.5 & 83.44 & 0.098 & 89.77 & 70.30 & 77.35 & 0.222 \\
Dolphin-1.5 & 83.38 & 0.106 & 89.30 & 71.43 & 75.86 & 0.215 \\
Mistral-OCR & 82.85 & 0.104 & 90.62 & 68.36 & 73.08 & 0.219 \\
OpenDoc-0.1B & 82.69 & 0.100 & 90.73 & 67.32 & 72.55 & 0.206 \\
HunyuanOCR & 82.69 & 0.120 & 80.32 & 79.76 & 84.92 & 0.243 \\
Nanonets-OCR-s & 76.90 & 0.154 & 70.99 & 75.05 & 81.56 & 0.309 \\
POINTS-Reader & 74.86 & 0.103 & 75.23 & 59.60 & 64.19 & 0.263 \\
\bottomrule
\end{tabular}
\end{table*}

Rankings on the Hard subset differ markedly from Base, validating its effectiveness in differentiating model capabilities. Key observations: (1) \minerupro leads at 94.08, ahead of the second-place PaddleOCR-VL (92.48) by 1.60 points, achieving the best scores in Formula CDM (97.54), Table TEDS (89.91), and Reading Order (0.170). (2) GLM-OCR ranks first on Base (96.19) but drops to third on Hard (92.01), a decline of 4.18 points; HunyuanOCR drops from Base (92.45) to Hard (82.69), a decline of 9.76 points. In contrast, \minerupro declines by only 2.04 points, demonstrating the strongest robustness. (3) Gemini 3 Pro/Flash perform well on Hard (91.99), narrowing the gap with specialized models, benefiting from their strong capabilities on hard formulas (CDM 97.23/96.68).

\section{Qualitative Comparison with SOTA Methods}
\label{app:qualitative_comparison}

This section presents qualitative comparisons of parsing results between \minerupro and current SOTA methods on representative scenarios.

\subsection{Table Recognition}
\label{app:qual_table}

\begin{figure*}[t]
\centering
\includegraphics[width=0.92\textwidth]{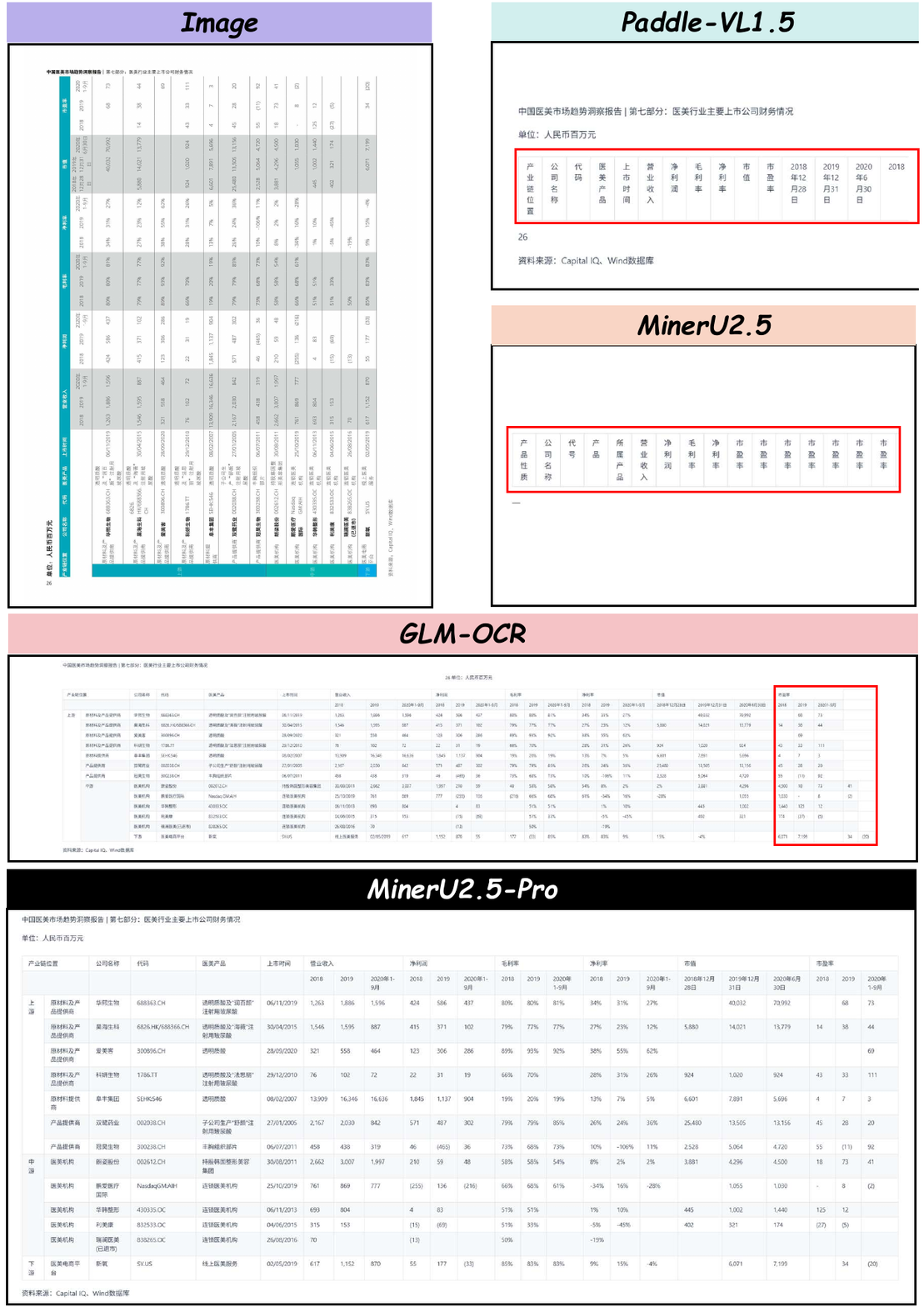}
\caption{Qualitative comparison on rotated table recognition. \minerupro correctly recovers the rotated structure, while competing models produce misaligned rows or missing cells.}
\label{fig:compare_table1}
\end{figure*}

\begin{figure*}[t]
\centering
\includegraphics[width=0.92\textwidth]{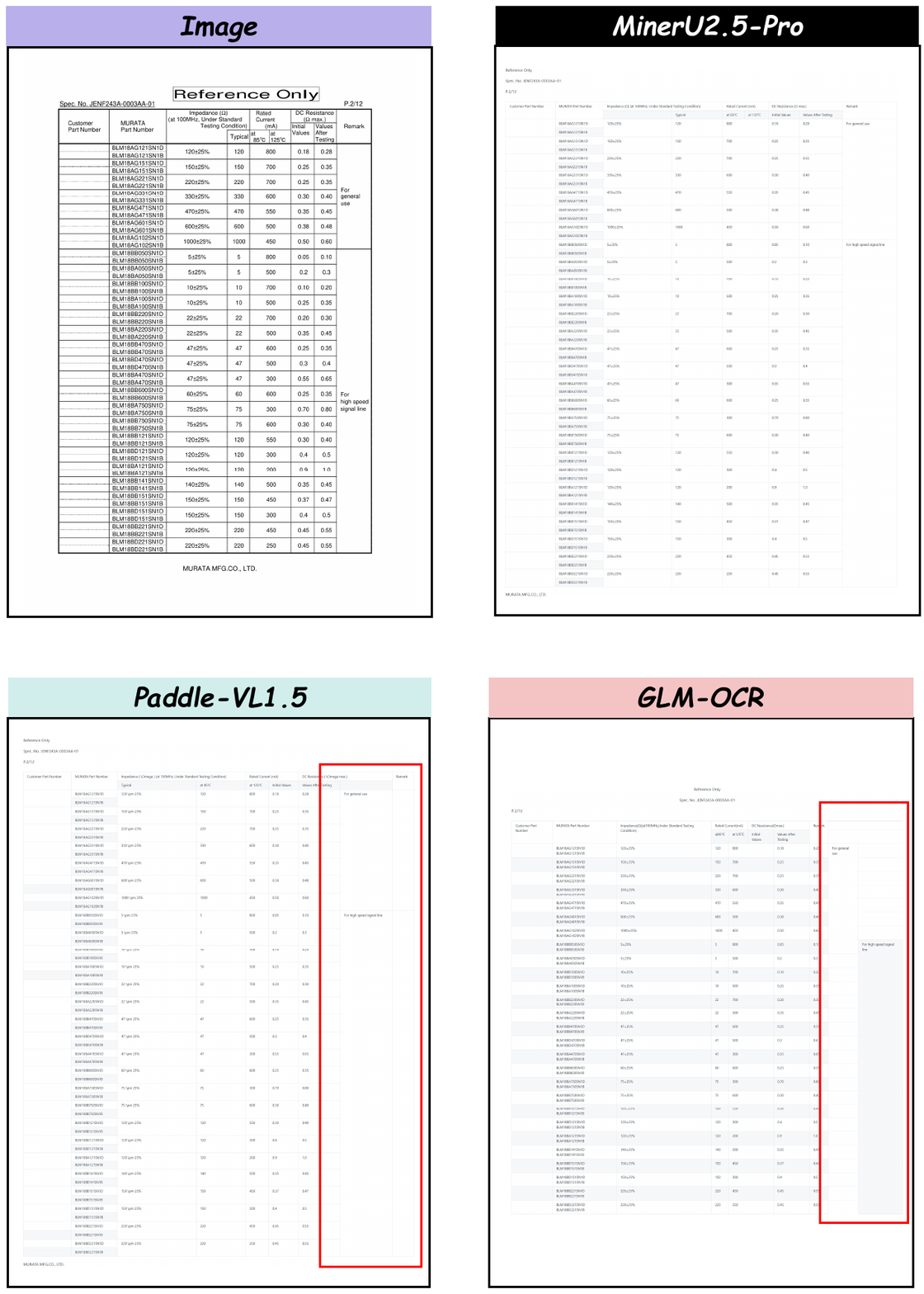}
\caption{Qualitative comparison on tables with long merged cells. \minerupro preserves the span structure, whereas other models incorrectly split or duplicate merged cells.}
\label{fig:compare_table2}
\end{figure*}

\minerupro demonstrates superior accuracy on complex tables, particularly rotated tables and tables with long merged cells. As shown in \Cref{fig:compare_table1} and \Cref{fig:compare_table2}, \minerupro correctly recovers the table structure and content, while competing models exhibit noticeable structural errors such as misaligned rows and lost cell boundaries.

\subsection{Formula Recognition}
\label{app:qual_formula}

\begin{figure*}[t]
\centering
\includegraphics[width=0.92\textwidth]{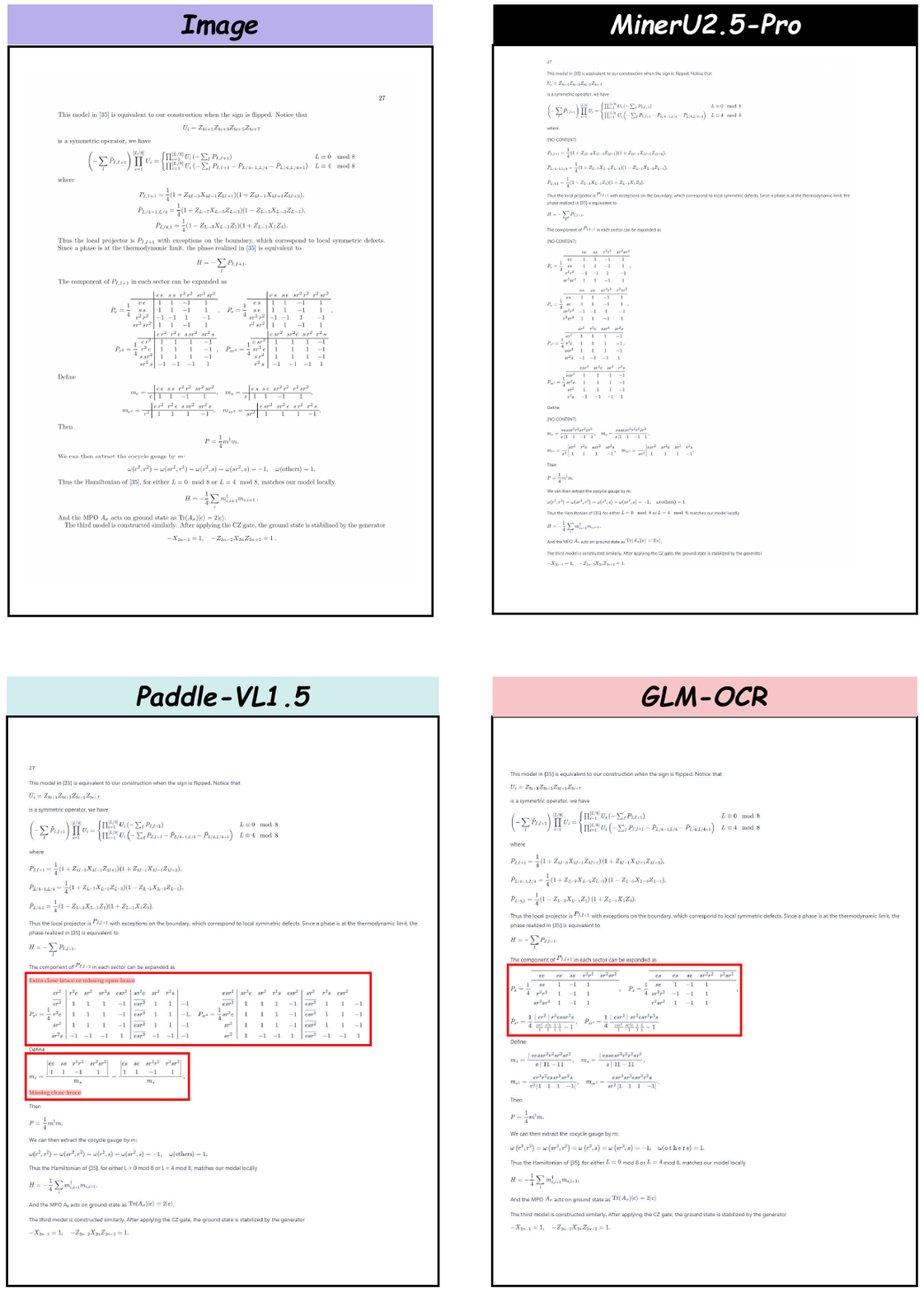}
\caption{Qualitative comparison on complex matrix recognition. \minerupro accurately captures the matrix structure and alignment, while other models exhibit symbol errors or structural collapse.}
\label{fig:compare_formula1}
\end{figure*}

\begin{figure*}[t]
\centering
\includegraphics[width=0.92\textwidth]{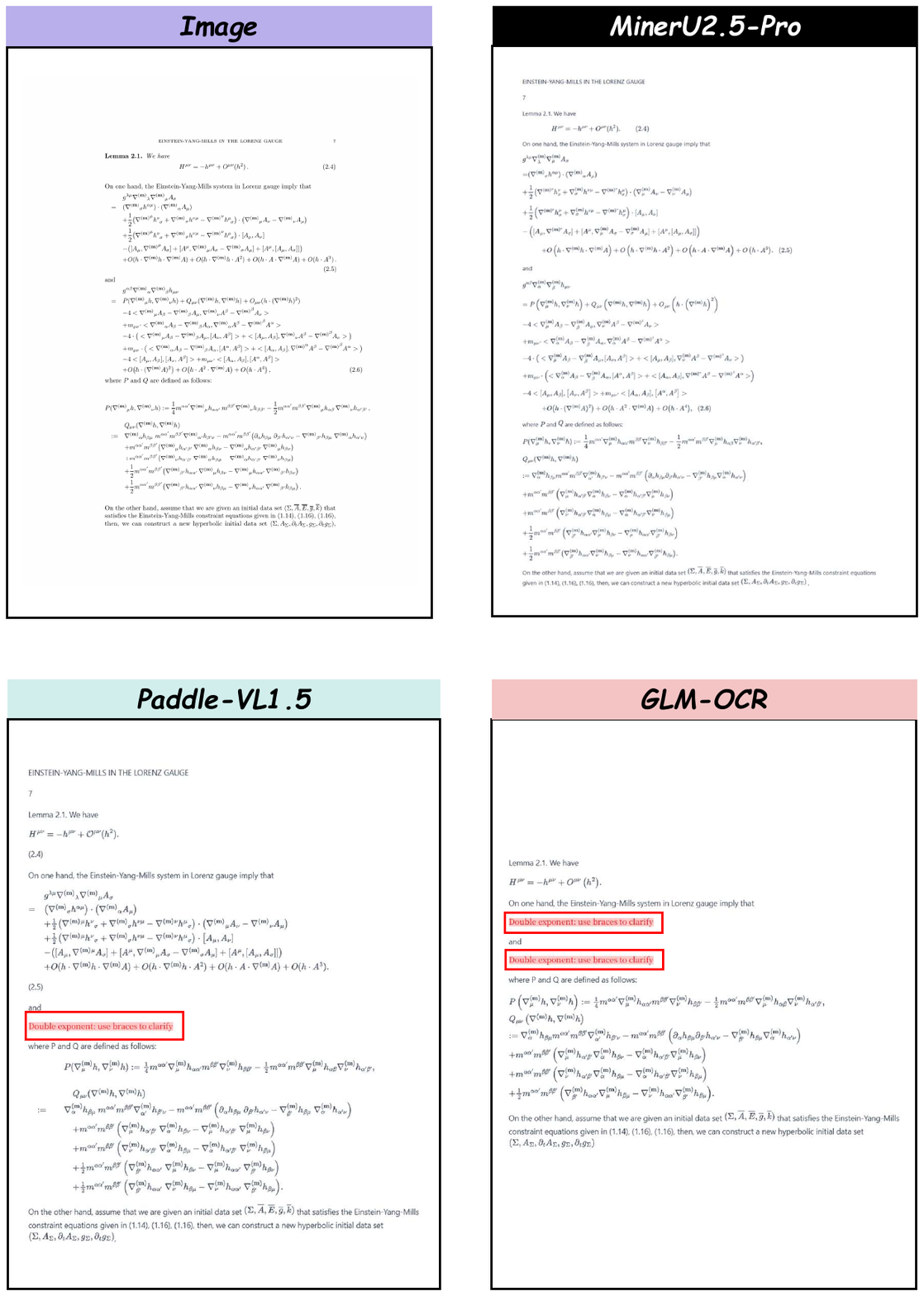}
\caption{Qualitative comparison on multi-line formula recognition. The row-by-row analysis of \minerupro faithfully reproduces each line, whereas competing models merge or misalign lines.}
\label{fig:compare_formula2}
\end{figure*}

The decoupled row-by-row formula analysis of \minerupro yields high accuracy on multi-line formulas, substantially outperforming end-to-end approaches that must generate entire equation groups in a single pass. \minerupro also achieves more accurate recognition on complex matrices. Representative comparisons are shown in \Cref{fig:compare_formula1} and \Cref{fig:compare_formula2}.

\subsection{Image-Aware Parsing}
\label{app:qual_text}

\begin{figure*}[t]
\centering
\includegraphics[width=0.92\textwidth]{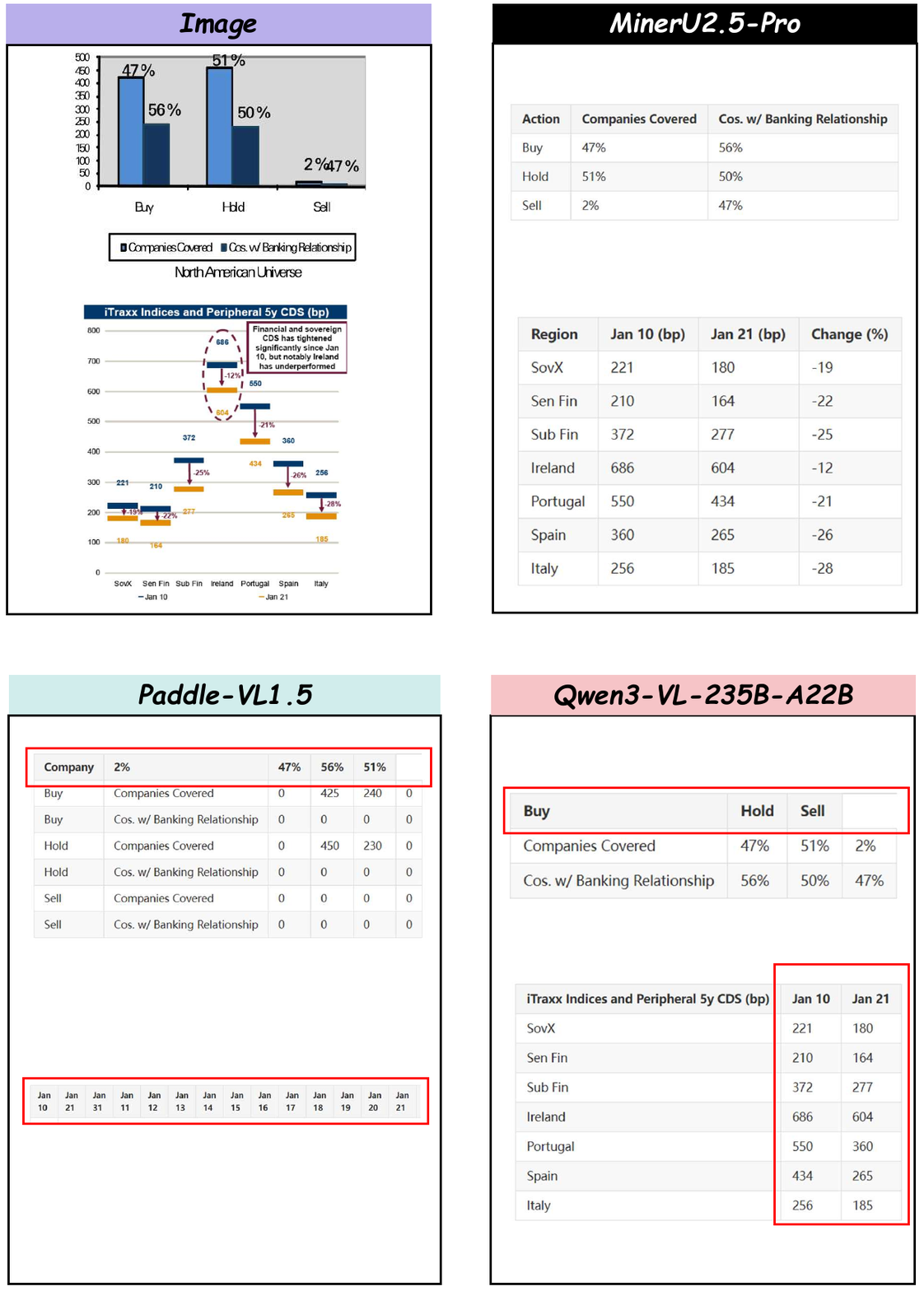}
\caption{Qualitative comparison on image-aware chart parsing (Part 1). \minerupro extracts structured content from diverse chart types, while other models either ignore or misinterpret chart content.}
\label{fig:compare_image1}
\end{figure*}

\begin{figure*}[t]
\centering
\includegraphics[width=0.92\textwidth]{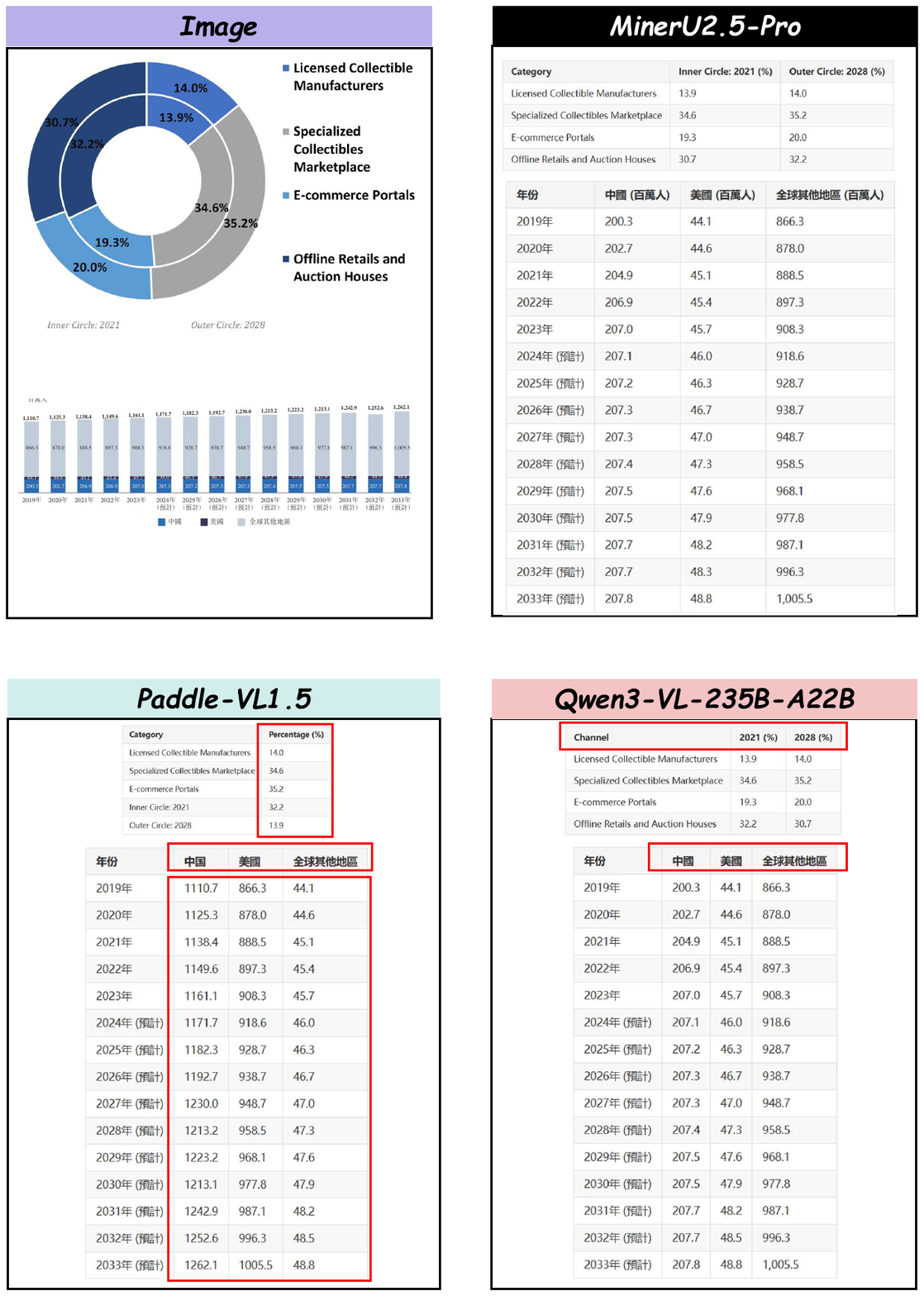}
\caption{Qualitative comparison on image-aware chart parsing (Part 2). Additional chart types demonstrating the generalization of \minerupro's image analysis pipeline.}
\label{fig:compare_image2}
\end{figure*}

\minerupro's image-aware parsing extracts structured content from chart and diagram regions that other models typically leave as opaque image placeholders. \Cref{fig:compare_image1} and \Cref{fig:compare_image2} compare parsing results across different chart types.

\end{document}